\documentclass[journal ]{new-aiaa}
\usepackage[utf8]{inputenc}
\usepackage{textcomp}

\usepackage{graphicx}
\usepackage{amsmath}
\usepackage[version=4]{mhchem}
\usepackage{siunitx}
\usepackage{longtable,tabularx}
\usepackage{verbatim}
\usepackage[ruled,vlined,linesnumbered,algo2e]{algorithm2e}
\usepackage{xcolor}
\usepackage{comment}
\DeclareMathOperator*{\argmin}{arg\,min}
\newcommand{\norm}[1]{\left\lVert#1\right\rVert}
\newcommand{\revision}[1]{{#1}} 

\title{A Closed-Loop Linear Covariance Framework for Vehicle Path Planning in a Static Uncertain Obstacle Field}

\author{Randall S Christensen \footnote{Corresponding Author, Assistant Professor, Department of Electrical and Computer Engineering, Randall.Christensen@usu.edu} 
and Greg Droge\footnote{Assistant Professor, Department of Electrical and Computer Engineering, Greg.Droge@usu.edu}}
\affil{Utah State University, Logan, Utah, 84322, USA}
\author{Robert C. Leishman\footnote{Director, Autonomy and Navigation Technology Center, Robert.Leishman@afit.edu}}
\affil{Air Force Institute of Technology, Wright-Patterson AFB, OH, 45433, USA}

\begin{document}

\maketitle

\begin{abstract}
Path planning in an uncertain environment is a key enabler of true vehicle autonomy. Over the past two decades, numerous approaches have been developed to account for errors in the vehicle path while navigating complex and often uncertain environments. An important capability of such planning is the prediction of vehicle dispersion covariances about a candidate path. This work develops a new closed-loop linear covariance (CL-LinCov) framework applicable to a wide range of autonomous system architectures.  Important features of the developed framework include the (1) separation of high-level guidance from low-level control, (2) support for output-feedback controllers with internal states, dynamics, and output, and (3) multi-use continuous sensors for navigation state propagation, guidance, and feedback control.  The closed-loop nature of the framework preserves the important coupling between the system dynamics, exogenous disturbances, and the guidance, navigation, and control algorithms. The developed framework is applied to a simplified model of an unmanned aerial vehicle and validated by comparison via Monte Carlo analysis. The utility of the CL-LinCov information is illustrated by its application to path planning in a \revision{static}, uncertain obstacle field via a modified version of the Rapidly Exploring Random Tree algorithm.


\end{abstract}

\section*{Nomenclature}
{\renewcommand\arraystretch{1.0}
\noindent\begin{longtable*}{@{}l @{\quad=\quad} l@{}}
$\boldsymbol{x}$ & truth states\\
$\boldsymbol{u}$ & continuous actuator value\\
$\hat{\boldsymbol{x}}$ & navigation state\\
$\tilde{\boldsymbol{y}}$ & continuous measurements\\
$\tilde{\boldsymbol{z}}_k$ & discrete measurements\\
$\hat{P}$ & navigation state error covariance\\
$P_{true}$ & true navigation state error covariance\\
$D_{true}$ & true state dispersion covariance\\
$\boldsymbol{x}_n$ & true value of navigation state\\
$\hat{\boldsymbol{x}}^*$ & desired state\\
$\check{\boldsymbol{x}}$ & control state\\
$\delta\boldsymbol{x}$ & true state dispersions\\
$\delta\hat{\boldsymbol{x}}$ & navigation state dispersions\\
$\delta\check{\boldsymbol{x}}$ & controller state dispersions\\
$\delta\boldsymbol{e}$ & true navigation error\\
$p_n$ & north position\\
$p_e$ & east position\\
$V_g$ & ground speed\\
$\psi$ & heading\\
$\omega$ & yaw rate\\
$u_w$ & axial wind gust\\
$T_{dist}$ & disturbance torque\\
$F_c$ & control force\\
$T_c$ & control torque\\
$\rho$ & air density\\
$C_{D_{0}}$ & drag coefficient\\
$S_p$ & planform area\\
$m$ & vehicle mass\\
$J$ & vehicle inertia\\
$L_u$ & wind gust correlation distance\\
$\boldsymbol{w}_i$ & waypoint location\\
$\mathcal{O}$ & obstacle map\\
$V$ & graph vertex\\
$E$ & graph edge
\end{longtable*}}
\section{Introduction}
\lettrine{A}{utonomous} vehicles are quickly becoming integral parts of both civilian and military systems. An important component of such systems is the ability to safely plan the path of the vehicle to achieve mission objectives. Robust path planning must account for the dispersion of the vehicle about a candidate path as well as the uncertainty in obstacles and/or adversarial systems.  An important aspect of the dispersions is the coupled nature of the vehicle dynamics and the guidance, navigation, and control (GNC) system.  Neglecting this coupling often leads to risk probabilities that are overly optimistic or altogether incorrect~\cite{bry_rapidly-exploring_2011}.  Monte Carlo analysis is the status-quo method for assessing uncertainty in complex GNC systems. Given a large number of simulations (500+), the computed statistics are a good representation of overall system performance. In the case of vehicle path planning, however, the number of candidate paths considered is often very large, precluding the use of Monte Carlo analysis for mission planning in general and especially for real-time mission planning or situations with large numbers of cooperating vehicles. Thus, there is a need for a broadly applicable covariance framework that produces dispersions statistics in a manner that is efficient for these more demanding scenarios.

Closed-loop Linear Covariance (CL-LinCov) analysis provides an efficient means for predicting the covariance of vehicle dispersions along a nominal trajectory. Under assumptions of Gaussian error sources and locally-linear dynamics and GNC algorithms, CL-LinCov produces the same statistical information as Monte Carlo analysis in a single run~\cite{christensen_linear_2014}.  This work utilizes CL-LinCov to plan the path of an autonomous vehicle through \revision{a static obstacle field, whose obstacle locations are poorly known}.  A new CL-LinCov framework is developed, applicable to a wide range of autonomous vehicles, then validated for the specific case of a Dubins vehicle with drag, gust, and torque disturbances.   Once validated, the dispersion covariance from the CL-LinCov framework is combined with obstacle location uncertainty to compute the probability of collision along candidate paths of a Rapidly-exploring Random Tree (RRT). 


The remainder of this paper is organized as follows.  A review of relevant literature is  provided in Section~\ref{sec:Literature-Review}.  Section~\ref{sec:General-Framework} defines the Monte Carlo and LinCov frameworks with sufficient complexity to model a wide variety of autonomous vehicles. Section~\ref{subsec:Vehicle-Model} defines the dynamics and GNC models for the vehicle considered in this research. Section~\ref{subsec:Path-Planning} discusses the modeling of \revision{static}, uncertain obstacles and the augmentation of the RRT algorithm. Section~\ref{sec:Simulation-Results} presents simulation results, which validate the LinCov framework and demonstrate its use for vehicle path planning.  Conclusions are summarized in Section~\ref{sec:Conclusion} with lengthy derivations included in the appendix.

%

\section{Literature Review\label{sec:Literature-Review}}
\begin{figure}
\begin{centering}
\includegraphics[viewport=245bp 195bp 605bp 385bp,clip,width=3.0in]{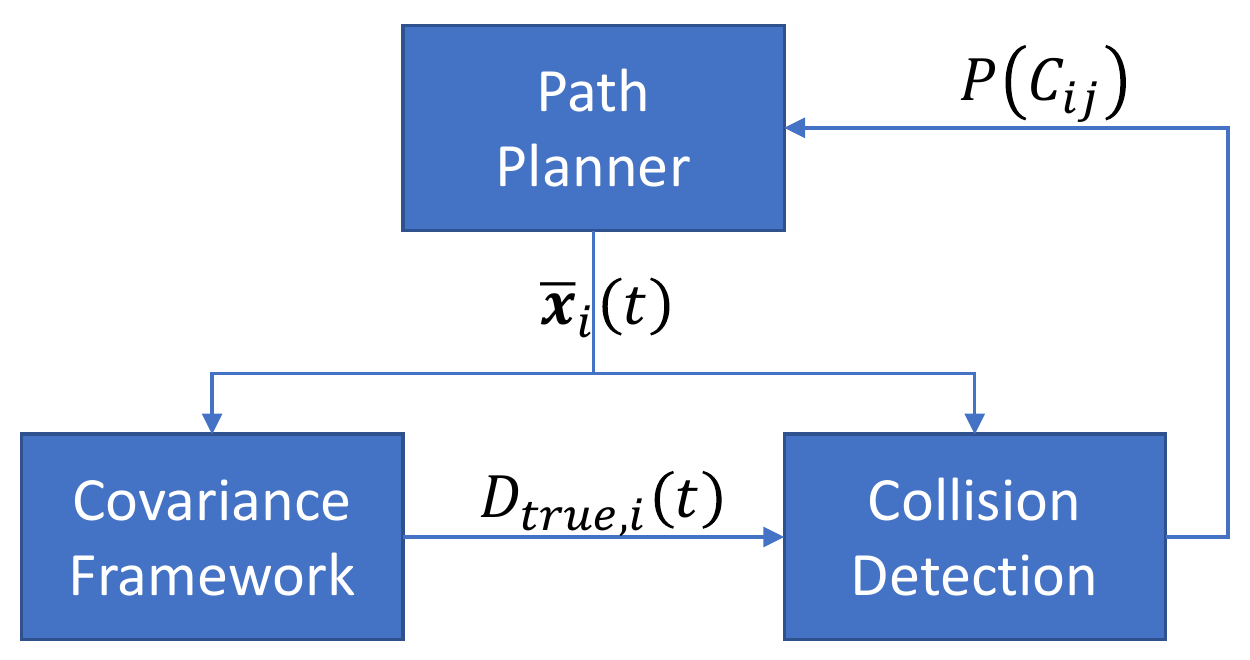}
\par\end{centering}
\caption{Path Planning With Uncertainty\label{fig:Path-Planning-In}}
\end{figure}
\revision{The concept of accounting for uncertainty during path planning is not new, and excellent surveys are available on the topic~\cite{dadkhah_survey_2012,hoy_algorithms_2015}.  A common paradigm is illustrated in Fig.~\ref{fig:Path-Planning-In}.  Here, a path planner produces a kinematically feasible candidate trajectory $\bar{\boldsymbol{x}}_i\left(t\right)$.  The covariance framework evaluates the uncertainty $D_{true,i}\left(t\right)$ of the vehicle dispersion about the candidate path.  Since uncertainty exists in the vehicle trajectory and potentially in the obstacle location, a deterministic evaluation of collision is not possible.  Rather, the probability of collision with the $j$th obstacle $P\left(C_{ij}\right)$ is evaluated and provided to the path planner.   The candidate path is then accepted or rejected based on a user-specified constraint for the probability of a collision, the so-called chance constraint~\cite{blackmore_chance-constrained_2011}.}

\revision{While some path planning frameworks have been developed to handle non-Gaussian statistics~\cite{shetty_trajectory_2020,summers_distributionally_2018}, the overwhelming majority assume Gaussian distributions for the vehicle state and obstacle position and a linear or linearized system ~\cite{berning_rapid_2020,bry_rapidly-exploring_2011,park_efficient_2017,park_fast_2016,van_den_berg_lqg-mp_2011,okamoto_optimal_2019,zhu_chance-constrained_2019, du_toit_probabilistic_2011,pairet_uncertainty-based_2018,watanabe_navigation_2016,castillo-lopez_real-time_2020,blackmore_chance-constrained_2011}.  Although the methods for computing the chance constraint differ, a fundamental quantity needed in the computations is the covariance matrix of the vehicle state about the candidate trajectory.  An increase in the generality of covariance frameworks for unmanned vehicles is the topic of this paper.}

\revision{In the context of linear, Gaussian covariance frameworks, a wide range of detail and fidelity is observed, with overlapping capabilities in modeling system architectures typical of unmanned systems.  A representative summary of the literature is provided in Table~\ref{tab:Summary-of-Covariance}, with the current work illustrated in the final column.  In all frameworks, a truth state differential equation is included, with random process noise to capture the effects of unknown, exogenous disturbances or modeling errors.  Higher fidelity frameworks include estimated states (i.e., the navigation states), discrete measurements, and an associated Kalman filter.  It is common to also include a full or partial state feedback control law, which uses knowledge of the truth or navigation states to regulate deviations from the candidate path. The technique of model replacement is less common, where navigation states are propagated using continuous sensors (e.g., accelerometer and gyros) rather than a kinetic model~\cite{maybeck_stochastic_1994}.  Less common still is the inclusion of an output feedback controller and associated controller state dynamics, which first appeared in recent research by the authors~\cite{christensen_closed-loop_2018}.  Also less common is the inclusion of a separated guidance law, similar to the architecture presented in~\cite{watanabe_navigation_2016}.  The term separated is used to differentiate such guidance laws from the combined guidance and feedback control modeled in other frameworks~\cite{geller_linear_2006,woffinden_linear_2019}.  A separated guidance law becomes important when considering cascaded autopilots common in UAVs~\cite{beard_randy_small_2012}, where low-level feedback control laws are driven by high-level guidance laws.}

\revision{Another important feature for autonomous vehicles is the continuous guidance feedback observed in the wall-following guidance law of~\cite{watanabe_navigation_2016}.  In this case, the guidance law operates directly on measurements of optical flow and range relative to a nearby wall, rather than on the navigation states alone. Lastly, one feature not observed in current frameworks is accommodation of continuous sensors that are used both as model replacement and in the feedback control.  This type of GNC architecture is commonly encountered in unmanned systems, e.g., where gyros are used to propagate attitude states and are also used as feedback in the rate damping loops~\cite{beard_randy_small_2012}.  While many of the desired features are available in the literature, an inclusive covariance framework is lacking.}


\begin{table}
\small
\begin{centering}
\begin{tabular}{lccccccccccc}
\hline
 Feature & \cite{berning_rapid_2020,okamoto_optimal_2019} & \cite{bry_rapidly-exploring_2011,van_den_berg_lqg-mp_2011} & \cite{zhu_chance-constrained_2019} & \cite{pairet_uncertainty-based_2018} & \cite{watanabe_navigation_2016} & \cite{castillo-lopez_real-time_2020,blackmore_chance-constrained_2011} & \cite{christensen_linear_2020} & \cite{geller_linear_2006,woffinden_linear_2019} & \cite{christensen_terrain-relative_2011} & \cite{christensen_closed-loop_2018} &\tabularnewline
 \hline
Truth state dynamics & x & x & x & x & x & x & x & x & x & x & x\tabularnewline
Truth process noise & x & x & x & x & x & x & x & x & x & x & x\tabularnewline
Navigation state dynamics &  & x &  & x & x &  & x & x & x & x & x\tabularnewline
Discrete measurements &  & x &  &  & x &  & x & x & x & x & x\tabularnewline
Kalman filter &  & x &  &  & x &  & x & x & x & x & x\tabularnewline
Model replacement &  &  &  &  & x &  & x & x & x & x & x\tabularnewline
Controller states &  &  &  &  &  &  &  &  &  & x & x\tabularnewline
State Feedback Control & x & x &  &  & x &  &  & x &  & x & x\tabularnewline
Output Feedback Control &  &  &  &  &  &  &  &  &  & x & x\tabularnewline
Separated Guidance &  &  &  &  & x &  &  &  &  &  & x\tabularnewline
Continuous Guidance Feedback &  &  &  &  & x &  &  &  &  &  & x\tabularnewline
Multi-Use Continuous Sensors &  &  &  &  &  &  &  &  &  &  & x\tabularnewline
\hline 
\end{tabular}
\par\end{centering}
\caption{Summary of Covariance Framework Features\label{tab:Summary-of-Covariance}}
\end{table}

\subsection{Contributions}
\revision{The primary contribution of this paper is the unification and extension of existing covariance frameworks to support a wider range of autonomous vehicles.  A new CL-LinCov framework is developed that combines the features of~\cite{watanabe_navigation_2016} and~\cite{christensen_closed-loop_2018}, and further extends the modeling capabilities of the combined framework.  In addition to the features common to both, the new framework includes the ability to model output feedback controllers, with associated controller states, while separating the guidance laws from the feedback control.  This combination accommodates the tiered GNC architectures common to many autonomous vehicles.  The guidance algorithms are designed to accommodate both state feedback and continuous sensor feedback, enabling guidance laws that operate independent of navigation states.  Lastly, the combined framework is extended to handle the case of multi-use continuous sensors, used both for navigation state propagation and feedback control. The new framework is validated via Monte Carlo analysis for a simplified UAV model.  The secondary contribution of this research is utilization of the developed framework for path planning in a static, uncertain obstacle field, where the \textit{CollisionFree} primitive of the Rapidly-Exploring Random Tree is modified to accommodate uncertainty in both the vehicle and obstacle location.}


\section{General Framework\label{sec:General-Framework}}
This section develops the general framework for a closed-loop GNC system of sufficient complexity to model the vehicle outlined in Section \ref{subsec:Vehicle-Model}. \revision{In the development that follows, nonlinear models for the dynamics, sensors, navigation, guidance and control systems are defined.  The nonlinear models are then linearized about a reference trajectory to yield a linear system of truth state, navigation state, and controller state dispersions.  The dispersions are then augmented to yield the desired CL-LinCov architecture.}


\subsection{Nonlinear Models\label{subsec:Nonlinear-Truth-Model}}
The general form of the nonlinear equations that define the dynamics, sensors, and GNC system are developed here. The truth dynamics evolve based on the truth states of the system $\boldsymbol{x}\in\mathbb{R}^{n}$, the input $\boldsymbol{u}\in\mathbb{R}^{n_{u}}$, and white noise $\boldsymbol{w}\in\mathbb{R}^{n_{w}}$ with power spectral density (PSD) $S_{w}$ according to
\begin{equation}
\dot{\boldsymbol{x}}=\boldsymbol{f}\left(\boldsymbol{x},\boldsymbol{u},\boldsymbol{w}\right)\label{eq:nonlin1}
\end{equation}

Observations of the truth state of the system are made through measurements, both in continuous and discrete time. Continuous measurements $\tilde{\boldsymbol{y}}\in\mathbb{R}^{n_{\tilde{y}}}$ 
are corrupted by continuous white noise $\boldsymbol{\eta}$ with PSD $S_{\eta}$. Similarly, discrete measurements $\tilde{\boldsymbol{z}}\in\mathbb{R}^{n_{z}}$ are corrupted by discrete noise $\boldsymbol{\nu}_{k}$ with variance $R_{\nu}$, i.e.,
\begin{equation}
\tilde{\boldsymbol{y}}=\boldsymbol{c}\left(\boldsymbol{x},\boldsymbol{u}\right)+\boldsymbol{\eta}\label{eq:nonlin2}
\end{equation}
\begin{equation}
\tilde{\boldsymbol{z}}_{k}=\boldsymbol{h}\left(\boldsymbol{x}_{k}\right)+\boldsymbol{\nu}_{k}\label{eq:nonlin4}
\end{equation}

The navigation state $\hat{\boldsymbol{x}}\in\mathbb{R}^{\hat{n}}$ is propagated using the continuous inertial measurements.  The navigation state is updated via the Kalman gain $\hat{K}_{k}$ and the residual $\tilde{\boldsymbol{z}}_{k}-\hat{\tilde{\boldsymbol{z}}}_{k}$, the difference between the discrete measurement and the predicted discrete measurement $\hat{\tilde{\boldsymbol{z}}}\in\mathbb{R}^{n_{z}}$, i.e.,
\begin{equation}
\dot{\hat{\boldsymbol{x}}}=\hat{\boldsymbol{f}}\left(\hat{\boldsymbol{x}},\tilde{\boldsymbol{y}}\right)\label{eq:nonlin5}
\end{equation}
\begin{equation}
\hat{\boldsymbol{x}}_{k}^{+}=\hat{\boldsymbol{x}}_{k}^{-}+\hat{K}_{k}\left[\tilde{\boldsymbol{z}}_{k}-\hat{\tilde{\boldsymbol{z}}}_{k}\right]\label{eq:nonlin6}
\end{equation}
\begin{equation}
\hat{\tilde{\boldsymbol{z}}}_{k}=\hat{\boldsymbol{h}}\left(\hat{\boldsymbol{x}}_{k}\right)\label{eq:nonlin7}
\end{equation}
\begin{equation}
\hat{K}_{k}=\hat{P}_{k}^{-}\hat{H}_{k}\left(\hat{H}_{k}\hat{P}_{k}^{-}\hat{H}_{k}^{T}+\hat{R}_{\nu}\right)^{-1}\label{eq:nonlin8}
\end{equation}

The navigation state covariance $\hat{P}$ is propagated via the continuous Ricatti equation and updated using the Joseph form
\begin{equation}
\dot{\hat{P}}=\hat{F}\hat{P}+\hat{P}\hat{F}^{T}+\hat{B}\hat{Q}\hat{B}^{T}
\end{equation}
\begin{equation}
\hat{P}_{k}^{+}=\left(I-\hat{K}_{k}\hat{H}_{k}\right)\hat{P}_{k}^{-}\left(I-\hat{K}_{k}\hat{H}_{k}\right)^{T}+\hat{K}_{k}\hat{R}_{\nu}\hat{K}_{k}^{T}\label{eq:nonlin10}
\end{equation}
The quantities $\hat{\boldsymbol{f}}$, $\hat{\boldsymbol{h}}$,
$\hat{F}$, $\hat{H}$, $\hat{Q}$, and $\hat{R}_{\nu}$ are all obtained
from the navigation design model and are allowed to differ from the true values.

In many problems, the truth state vector is a more comprehensive description of the system and is therefore larger and possibly of a different form than the navigation state vector. The true values of the navigation state $\boldsymbol{x}_n$ are given by the following mapping function of the truth states:
\begin{equation}
\boldsymbol{x}_{n}=\boldsymbol{m}\left(\boldsymbol{x}\right)\label{eq:nonlin11}
\end{equation}

The guidance law is modeled as a potentially nonlinear function of the navigation state \revision{and continuous measurements}, and produces a \revision{desired state $\hat{\boldsymbol{x}}^{*}\in\mathbb{R}^{\hat{n}^{*}}$} given by
\begin{equation}
\hat{\boldsymbol{x}}^{*}=\boldsymbol{n}\left(\hat{\boldsymbol{x}},\tilde{\boldsymbol{y}}\right)\label{eq:nonlin12}
\end{equation}
Note that the guidance algorithm does not, in general, specify the desired value for all components of the state vector. The output $\hat{\boldsymbol{x}}^{*}$ is therefore of size $\hat{n}^{*}$, which is not necessarily equal to $\hat{n}$. To enable modeling of output feedback control laws (e.g., PID controllers), a control state $\check{\boldsymbol{x}}\in\mathbb{R}^{\hat{n}}$ is defined with a dynamics equation $\check{\boldsymbol{f}}$ and an output equation \revision{$\boldsymbol{g}\in\mathbb{R}^{n_{u}}$}, i.e.,
\begin{equation}
\dot{\check{\boldsymbol{x}}}=\check{\boldsymbol{f}}\left(\hat{\boldsymbol{x}},\hat{\boldsymbol{x}}^{*}\right)\label{eq:nonlin13}
\end{equation}
\begin{equation}
\boldsymbol{u}=\boldsymbol{g}\left(\check{\boldsymbol{x}},\hat{\boldsymbol{x}},\hat{\boldsymbol{x}}^{*},\tilde{\boldsymbol{y}}\right)\label{eq:nonlin14}
\end{equation}
\revision{Note that the developed CL-LinCov framework provides increased flexibility in the use of the continuous measurement vector $\tilde{\boldsymbol{y}}$  defined in Eq.~\eqref{eq:nonlin5}. Some or potentially all components are include in the guidance law of Eq.~\eqref{eq:nonlin14} and feedback control law of Eq.~\eqref{eq:nonlin12}.}

\subsection{Linear Modeling\label{subsec:Linear-Modeling}}
To enable efficient covariance calculations, this section linearizes the system in Section~\ref{subsec:Nonlinear-Truth-Model} about a nominal reference trajectory.  An important concept in the linearization is the dispersion of the truth $\boldsymbol{x}$, navigation $\hat{\boldsymbol{x}}$, and controller states $\check{\boldsymbol{x}}$. The dispersions are defined as the difference between the state and its corresponding nominal, denoted by an overbar, i.e.,
\begin{equation}
\delta\boldsymbol{x}=\boldsymbol{x}-\bar{\boldsymbol{x}}
\end{equation}
\begin{equation}
\delta\hat{\boldsymbol{x}}=\hat{\boldsymbol{x}}-\hat{\bar{\boldsymbol{x}}}
\end{equation}
\begin{equation}
\delta\check{\boldsymbol{x}}=\check{\boldsymbol{x}}-\check{\bar{\boldsymbol{x}}}
\end{equation}

There are multiple ways, in practice, to compute the nominal trajectory. For
this research, the nominal is \revision{generated by simulation of the system in Section~\ref{subsec:Nonlinear-Truth-Model}, without the uncertainty induced by process noise and measurement noise, i.e., $S_w=0_{n_w \times n_w} $, $S_\eta=0_{n_{\tilde{y}} \times n_{\tilde{y}}}$, and $R_\nu=0_{n_z \times n_z}$.} The remainder
of this section linearizes the system defined in Section \ref{subsec:Nonlinear-Truth-Model} about the nominal trajectory to obtain a coupled, linear state space system in the dispersion variables $\delta\boldsymbol{x}$, $\delta\hat{\boldsymbol{x}}$, and $\delta\check{\boldsymbol{x}}$. This step is performed via partial derivatives of the nonlinear function with respect one of the arguments. As an example of the notation, $F_{x}$ denotes the partial derivative of the vector function $\boldsymbol{f}$ with respect to the vector $\boldsymbol{x}$, evaluated along the nominal trajectory, as in
\begin{equation}
 F_{x}=\left.\frac{\partial\boldsymbol{f}}{\partial\boldsymbol{x}}\right|_{\boldsymbol{x}=\bar{\boldsymbol{x}},\,\hat{\boldsymbol{x}}=\hat{\bar{\boldsymbol{x}}},\,\check{\boldsymbol{x}}=\check{\bar{\boldsymbol{x}}}}   
\end{equation}
In the event that partial derivatives are difficult or impossible to derive, terms such as $F_{x}$ refer to the coefficient matrix obtained via alternative linearization techniques.

The differential equations describing the dynamics of the truth state Eq.~\eqref{eq:nonlin1}, navigation state Eq.~\eqref{eq:nonlin5}, and controller state \revision{Eq.~\eqref{eq:nonlin13}} are linearized to produce the following perturbation differential equations:
\begin{eqnarray}
\delta\dot{\boldsymbol{x}} & = & \left[F_{x}+F_{u}S\left(G_{\hat{x}^{*}}N_{\tilde{y}}C_{x}+G_{\tilde{y}}C_{x}\right)\right]\delta\boldsymbol{x}\nonumber \\
 & + & F_{u}S\left(G_{\hat{x}}+G_{\hat{x}^{*}}N_{\hat{x}}\right)\delta\hat{\boldsymbol{x}}\nonumber \\
 & + & F_{u}SG_{\check{x}}\delta\check{\boldsymbol{x}}\nonumber \\
 & + & F_{u}S\left(G_{\hat{x}^{*}}N_{\tilde{y}}+G_{\tilde{y}}\right)\boldsymbol{\eta}+B\boldsymbol{w}\label{eq:linmod4}
\end{eqnarray}
\begin{eqnarray}
\delta\dot{\hat{\boldsymbol{x}}} & = & \hat{F}_{\tilde{y}}TC_{x}\delta\boldsymbol{x}\nonumber \\
 & + & \left[\hat{F}_{\hat{x}}+\hat{F}_{\tilde{y}}TC_{u}\left(G_{\hat{x}}+G_{\hat{x}^{*}}N_{\hat{x}}\right)\right]\delta\hat{\boldsymbol{x}}\nonumber \\
 & + & \hat{F}_{\tilde{y}}TC_{u}G_{\check{x}}\delta\check{\boldsymbol{x}}+\hat{F}_{\tilde{y}}T\boldsymbol{\eta}\label{eq:linmod5}
\end{eqnarray}
\begin{eqnarray}
\delta\dot{\check{\boldsymbol{x}}} & = & \check{F}_{\hat{x}^{*}}N_{\tilde{y}}TC_{x}\delta\boldsymbol{x}\nonumber \\
 & + & \left[\check{F}_{\hat{x}}+\check{F}_{\hat{x}^{*}}N_{\hat{x}}+\check{F}_{\hat{x}^{*}}N_{\tilde{y}}TC_{u}\left(G_{\hat{x}}+G_{\hat{x}^{*}}N_{\hat{x}}\right)\right]\delta\hat{\boldsymbol{x}}\nonumber \\
 & + & \check{F}_{\hat{x}^{*}}N_{\tilde{y}}TC_{u}G_{\check{x}}\delta\check{\boldsymbol{x}}\nonumber \\
 & + & \check{F}_{\hat{x}^{*}}N_{\tilde{y}}T\boldsymbol{\eta}\label{eq:linmod6}
\end{eqnarray}
where
\begin{equation}
S=\left(I_{n_{u}\times n_{u}}-G_{\hat{x}^{*}}N_{\tilde{y}}C_{u}-G_{\tilde{y}}C_{u}\right)^{-1}
\end{equation}
\begin{equation}
T=\left(I_{n_{y}\times n_{y}}-C_{u}G_{\hat{x}^{*}}N_{\tilde{y}}-C_{u}G_{\tilde{y}}\right)^{-1}
\end{equation}

The navigation state update Eq.~\eqref{eq:nonlin6} is also linearized
along with Eq.~\eqref{eq:nonlin7} to produce the following set
of dispersion update equations:
\begin{equation}
\delta\boldsymbol{x}_{k}^{+}=\delta\boldsymbol{x}_{k}^{+}\label{eq:linmod7}
\end{equation}
\begin{equation}
\delta\hat{\boldsymbol{x}}_{k}^{+}=\left(I_{\hat{n}\times\hat{n}}-\hat{K}_{k}\hat{H}_{\hat{x}}\right)\delta\hat{\boldsymbol{x}}_{k}^{-}+\hat{K}_{k}H_{x}\delta\boldsymbol{x}_{k}^{-}+\hat{K}_{k}\boldsymbol{\nu}_{k}\label{eq:linmod8}
\end{equation}
\begin{equation}
\delta\check{\boldsymbol{x}}_{k}^{+}=\delta\check{\boldsymbol{x}}_{k}^{+}\label{eq:linmod9}
\end{equation}
where the truth and controller state dispersions
are not directly affected by the update of the navigation state. The
derivation of the preceding dispersion propagation and update equations are
documented in \revision{Appendix A}.

Once linearized, the augmented state vector is formed as
\begin{equation}
\boldsymbol{X}=\left[\begin{array}{c}
\delta\boldsymbol{x}\\
\delta\hat{\boldsymbol{x}}\\
\delta\check{\boldsymbol{x}}
\end{array}\right]
\end{equation}
With this definition, the augmented propagation and update equations become
\begin{equation}\label{eq:augmented-diff-eq}
\dot{\boldsymbol{X}}=\mathcal{F}\boldsymbol{X}+\mathcal{G}\boldsymbol{\eta}+\mathcal{W}\boldsymbol{w}
\end{equation}
\begin{equation}\label{eq:augmented-update}
\boldsymbol{X}_{k}^{+}=\mathcal{A}_{k}\boldsymbol{X}_{k}^{-}+\mathcal{B}_{k}\boldsymbol{\nu}_{k}
\end{equation}
where
\begin{equation}
\mathcal{F}=\left[\begin{array}{ccc}
\mathcal{F}_{xx} & \mathcal{F}_{x\hat{x}} & \mathcal{F}_{x\check{x}}\\
\mathcal{F}_{\hat{x}x} & \mathcal{F}_{\hat{x}\hat{x}} & \mathcal{F}_{\hat{x}\check{x}}\\
\mathcal{F}_{\check{x}x} & \mathcal{F}_{\check{x}\hat{x}} & \mathcal{F}_{\check{x}\check{x}}
\end{array}\right]
\end{equation}
\begin{equation}
\mathcal{G}=\left[\begin{array}{c}
F_{u}S\left(G_{\hat{x}^{*}}N_{\tilde{y}}+G_{\tilde{y}}\right)\\
\hat{F}_{\tilde{y}}T\\
\check{F}_{\hat{x}^{*}}N_{\tilde{y}}T
\end{array}\right]
\end{equation}
\begin{equation}
\mathcal{W}=\left[\begin{array}{c}
B\\
0_{\hat{n}\times n_{w}}\\
0_{\check{n}\times n_{w}}
\end{array}\right]
\end{equation}
\begin{equation}
\mathcal{A}_{k}=\left[\begin{array}{ccc}
I_{n\times n} & 0_{n\times\hat{n}} & 0_{n\times\check{n}}\\
\hat{K}_{k}H_{x} & I_{\hat{n}\times\hat{n}}-\hat{K}_{k}\hat{H}_{\hat{x}} & 0_{\hat{n}\times\check{n}}\\
0_{\check{n}\times n} & 0_{\check{n}\times\hat{n}} & I_{\check{n}\times\check{n}}
\end{array}\right]
\end{equation}
\begin{equation}
\mathcal{B}_{k}=\left[\begin{array}{c}
0_{n\times n_{z}}\\
\hat{K}_{k}\\
0_{\check{n}\times n_{z}}
\end{array}\right]
\end{equation}
The individual sub matrices of $\mathcal{F}$ are defined as
\begin{equation}
\mathcal{F}_{xx}=F_{x}+F_{u}S\left(G_{\hat{x}^{*}}N_{\tilde{y}}C_{x}+G_{\tilde{y}}C_{x}\right)
\end{equation}
\begin{equation}
\mathcal{F}_{x\hat{x}}=F_{u}S\left(G_{\hat{x}}+G_{\hat{x}^{*}}N_{\hat{x}}\right)
\end{equation}
\begin{equation}
\mathcal{F}_{x\check{x}}=F_{u}SG_{\check{x}}
\end{equation}
\begin{equation}
\mathcal{F}_{\hat{x}x}=\hat{F}_{\tilde{y}}TC_{x}
\end{equation}
\begin{equation}
\mathcal{F}_{\hat{x}\hat{x}}=\hat{F}_{\hat{x}}+\hat{F}_{\tilde{y}}TC_{u}G_{\hat{x}}+\hat{F}_{\tilde{y}}TC_{u}G_{\hat{x}^{*}}N_{\hat{x}}
\end{equation}
\begin{equation}
\mathcal{F}_{\hat{x}\check{x}}=\hat{F}_{\tilde{y}}TC_{u}G_{\check{x}}
\end{equation}
\begin{equation}
\mathcal{F}_{\check{x}x}=\check{F}_{\hat{x}^{*}}N_{\tilde{y}}TC_{x}
\end{equation}
\begin{equation}
\mathcal{F}_{\check{x}\hat{x}}=\check{F}_{\hat{x}}+\check{F}_{\hat{x}^{*}}N_{\hat{x}}+\check{F}_{\hat{x}^{*}}N_{\tilde{y}}T\left(C_{u}G_{\hat{x}}+C_{u}G_{\hat{x}^{*}}N_{\hat{x}}\right)
\end{equation}
\begin{equation}
\mathcal{F}_{\check{x}\check{x}}=\check{F}_{\hat{x}^{*}}N_{\tilde{y}}TC_{u}G_{\check{x}}
\end{equation}

\subsection{Covariance Evaluation \label{subsec:Performance-Evaluation}}

With the augmented differential and update equations defined in Eqs.~\eqref{eq:augmented-diff-eq} and~\eqref{eq:augmented-update}, the covariance propagation and update equations are expressed as
\begin{equation}
    \dot{C}_{A}=\mathcal{F}C_{A}+C_{A}\mathcal{F}^{T}+\mathcal{G}S_{\eta}\mathcal{G}^{T}+\mathcal{W}S_{w}\mathcal{W}^{T}
\end{equation}
\begin{equation}
    C_{A}^{+} = \mathcal{A}_{k}C_{A}^{-}\mathcal{A}_{k}^{T}+\mathcal{B}_{k}R_{\nu}\mathcal{B}_{k}^{T}
\end{equation}
Two quantities of typical interest are the covariance of the truth state dispersions and the estimation errors. It is important to note that the estimation error covariance defined below is the true estimation error covariance, which may be different than the estimated quantity $\hat{P}$ from the Kalman filter. These quantities are extracted from the augmented covariance matrix via the following equations:
\begin{equation}
D_{true}=\left[\begin{array}{ccc}
I_{n\times n} & 0_{n\times\hat{n}} & 0_{n\times\check{n}}\end{array}\right]C_{A}\left[\begin{array}{ccc}
I_{n\times n} & 0_{n\times\hat{n}} & 0_{n\times\check{n}}\end{array}\right]^{T}
\end{equation}
\begin{equation}
P_{true}=\left[\begin{array}{ccc}
-M_{x} & I_{\hat{n}\times\hat{n}} & 0_{\hat{n}\times\check{n}}\end{array}\right]C_{A}\left[\begin{array}{ccc}
-M_{x} & I_{\hat{n}\times\hat{n}} & 0_{\hat{n}\times\check{n}}\end{array}\right]^{T}
\end{equation}

\section{Vehicle Model\label{subsec:Vehicle-Model}}
This section of the paper defines the nonlinear equations specific to the UAV system considered in this research.  For demonstration purposes and mathematical tractability, this research utilizes the aircraft dynamics model found in~\cite{beard_randy_small_2012}, reduced to the horizontal plane and with assumptions of constant altitude and zero side-slip.  Despite the simplifications, the presented model has several key components of the more generic case of a full 6 degree-of-freedom model, and exercises the elements of the CL-LinCov framework developed in Section~\ref{sec:General-Framework}.  Where appropriate, the general equations from Section~\ref{sec:General-Framework} are indicated below the corresponding equation for the UAV system.

\subsection{Nonlinear Model}\label{sec:NonlinearVehicleModel}
The truth state of the vehicle in Eq.~\eqref{eq:truth-state-vector} is defined by its north and east position $(p_n, p_e)$, ground speed $(V_g)$, heading $(\psi)$, and angular rate $(\omega)$.  The vehicle is disturbed by axial wind gusts $(u_w)$ and disturbance torques $(T_d)$, which are also included in the truth state.
\begin{equation}
\label{eq:truth-state-vector}
\boldsymbol{x}=\left[\begin{array}{c}
p_{n}\\
p_{e}\\
V_{g}\\
\psi\\
\omega\\
u_{w}\\
T_{dist}
\end{array}\right]
\end{equation}

The dynamics of the truth states incorporate the kinematic relationship between position and velocity as well as heading and angular rate.  Aerodynamic effects include drag and a Dryden gust model~\cite{langelaan_wind_2011}, which is assumed to be predominantly in the \revision{direction of travel}.  Finally the disturbance torques are modeled as a First-Order Gauss-Markov (FOGM) process.  The truth state dynamics are
\begin{equation}
\underset{\dot{\boldsymbol{x}}}{\underbrace{\left[\begin{array}{c}
\dot{p}_{n}\\
\dot{p}_{e}\\
\dot{V}_{g}\\
\dot{\psi}\\
\dot{\omega}\\
\dot{u}_{w}\\
\dot{T}_{d}
\end{array}\right]}}=\underset{\boldsymbol{f}\left(\boldsymbol{x},\boldsymbol{u},\boldsymbol{w}\right)}{\underbrace{\left[\begin{array}{c}
V_{g}\cos\psi\\
V_{g}\sin\psi\\
\frac{1}{m}\left[F_{c}-\frac{1}{2}\rho C_{D_{0}}S_{p}\left(V_{g}-u_{w}\right)^{2}\right]\\
\omega\\
\frac{1}{J}\left(T_{c}+T_{dist}\right)\\
-\frac{V_{g}}{L_{u}}u_{w}+\sigma_{u}\sqrt{\frac{2V_{g}}{L_{u}}}w_{u}\\
-\frac{1}{\tau_{T}}T_{d}+w_{T}
\end{array}\right]}}\label{eq:vehicle1}
\end{equation}
The control inputs to the vehicle are force $F_c$ and torque $T_c$, i.e.,
\begin{equation}
\boldsymbol{u}=\left[\begin{array}{c}
F_{c}\\
T_{c}
\end{array}\right]
\end{equation}
Finally, the white process noise includes the driving noise for the Dryen gust model and the disturbance torque, i.e.,
\begin{equation}
\boldsymbol{w}=\left[\begin{array}{c}
w_{u}\\
w_{T}
\end{array}\right]
\end{equation}

The continuous measurements,
which are used to propagate the navigation state, consist of an accelerometer
aligned with the direction of travel, and a gyro aligned with the
local vertical, corrupted by continuous white noise, i.e.,
\begin{equation}
\underset{\tilde{\boldsymbol{y}}}{\underbrace{\left[\begin{array}{c}
\tilde{a}_{x}\\
\tilde{\omega}
\end{array}\right]}}=\underset{\boldsymbol{c}\left(\boldsymbol{x},\boldsymbol{u}\right)}{\underbrace{\left[\begin{array}{c}
a\\
\omega
\end{array}\right]}}+\underset{\boldsymbol{\eta}}{\underbrace{\left[\begin{array}{c}
\eta_{a}\\
\eta_{\omega}
\end{array}\right]}}\label{eq:vehicle6}
\end{equation}
where the true acceleration of the vehicle is a function of the control input $\boldsymbol{u}$
\begin{equation}
a=\frac{1}{m}\left\{ F_{control}-\frac{1}{2}\rho C_{D_{0}}S_{p}\left(V_{g}-u_{w}\right)^{2}\right\} 
\end{equation}
The discrete measurements available to the inertial navigation system are comprised of horizontal position and ground speed, corrupted by discrete white noise:
\begin{equation}
\underset{\tilde{\boldsymbol{z}}_{k}}{\underbrace{\left[\begin{array}{c}
\tilde{p}_{n}\left[t_{k}\right]\\
\tilde{p}_{e}\left[t_{k}\right]\\
\tilde{V}_{g}\left[t_{k}\right]
\end{array}\right]}}=\underset{\boldsymbol{h}\left(\boldsymbol{x}_{k}\right)}{\underbrace{\left[\begin{array}{c}
p_{n}\left[t_{k}\right]\\
p_{e}\left[t_{k}\right]\\
V_{g}\left[t_{k}\right]
\end{array}\right]}}+\underset{\boldsymbol{\nu}_{k}}{\underbrace{\left[\begin{array}{c}
\nu_{n}\left[t_{k}\right]\\
\nu_{e}\left[t_{k}\right]\\
\nu_{V}\left[t_{k}\right]
\end{array}\right]}}\label{eq:vehicle7}
\end{equation}

The navigation state vector is a subset of the truth states, including only position, ground speed, and heading states, i.e.,
\begin{equation}
\hat{\boldsymbol{x}}=\left[\begin{array}{c}
\hat{p}_{n}\\
\hat{p}_{e}\\
\hat{V}_{g}\\
\hat{\psi}
\end{array}\right]
\end{equation}
Since the vehicle has no knowledge of control and disturbance inputs, the navigation states are propagated using the so-called model replacement technique~\cite{maybeck_stochastic_1994}, where sensed accelerations and angular rates are utilized in lieu of the unknown inputs, i.e.,
\begin{equation}
\underset{\dot{\hat{\boldsymbol{x}}}}{\underbrace{\left[\begin{array}{c}
\dot{\hat{p}}_{n}\\
\dot{\hat{p}}_{e}\\
\dot{\hat{V}}_{g}\\
\dot{\hat{\psi}}
\end{array}\right]}}=\underset{\hat{\boldsymbol{f}}\left(\hat{\boldsymbol{x}},\tilde{\boldsymbol{y}}\right)}{\underbrace{\left[\begin{array}{c}
\hat{V}_{g}\cos\hat{\psi}\\
\hat{V}_{g}\sin\hat{\psi}\\
a+\eta_{a}\\
\omega+\eta_{\omega}
\end{array}\right]}}\label{eq:nav1}
\end{equation}
Finally, the predicted discrete measurement is equivalent to Eq.~\eqref{eq:vehicle7}, with the unknown measurement noise omitted, i.e.,
\begin{equation}
\underset{\hat{\tilde{\boldsymbol{z}}}}{\underbrace{\left[\begin{array}{c}
\hat{\tilde{p}}_{n}\left[t_{k}\right]\\
\hat{\tilde{p}}_{e}\left[t_{k}\right]\\
\hat{\tilde{V}}_{g}\left[t_{k}\right]
\end{array}\right]}}=\underset{\hat{\boldsymbol{h}}\left(\hat{\boldsymbol{x}}\right)}{\underbrace{\left[\begin{array}{c}
\hat{p}_{n}\left[t_{k}\right]\\
\hat{p}_{e}\left[t_{k}\right]\\
\hat{V}_{g}\left[t_{k}\right]
\end{array}\right]}}\label{eq:nav4}
\end{equation}

The matrices utilized in Eqs.~\eqref{eq:nonlin8} to~\eqref{eq:nonlin10} are derived by linearization of the preceding navigation models to obtain
\begin{equation}
\hat{F}=\left[\begin{array}{cccc}
0 & 0 & \cos\hat{\psi} & -\hat{V}_{g}\sin\hat{\psi}\\
0 & 0 & \sin\hat{\psi} & \hat{V}_{g}\cos\hat{\psi}\\
0 & 0 & 0 & 0\\
0 & 0 & 0 & 0
\end{array}\right]
\end{equation}
\begin{equation}
\hat{H}_{k}=\left[\begin{array}{cccc}
1 & 0 & 0 & 0\\
0 & 1 & 0 & 0\\
0 & 0 & 1 & 0
\end{array}\right]
\end{equation}
\begin{equation}
\hat{R}_{\nu}=\left[\begin{array}{ccc}
\hat{\sigma}_{n}^{2} & 0 & 0\\
0 & \hat{\sigma}_{e}^{2} & 0\\
0 & 0 & \hat{\sigma}_{v}^{2}
\end{array}\right]\delta_{ij}
\end{equation}
\begin{equation}
\hat{B}=\left[\begin{array}{cc}
0 & 0\\
0 & 0\\
1 & 0\\
0 & 1
\end{array}\right]
\end{equation}
\begin{equation}
\hat{Q}=\left[\begin{array}{cc}
\hat{Q}_{a} & 0\\
0 & \hat{Q}_{\omega}
\end{array}\right]
\end{equation}

It remains to define the guidance and low-level control laws of the UAV path following system. The guidance law is based on the straight-line path guidance presented in~\cite{beard_randy_small_2012}.  The output of the guidance laws comprises a constant commanded ground speed $V_g^*$ and a vehicle heading $\psi^*$ that directs the UAV to a straight line, i.e.,
\begin{equation}
\hat{\boldsymbol{x}}^{*}=\left[\begin{array}{c}
V_{g}^{*}\\
\psi^{*}
\end{array}\right]=\underset{\boldsymbol{n}\left(\hat{\boldsymbol{x}},\tilde{\boldsymbol{y}}\right)}{\underbrace{\left[\begin{array}{c}
\bar{V}_{g}\\
\psi_{q}-\psi^{\infty}\frac{2}{\pi}\arctan\left(k_{path}e_{path}\right)
\end{array}\right]}}\label{eq:guidance}
\end{equation}
where $\psi_{q}$ is the path heading, $e_{path}$ is the deviation of the vehicle from the desired path in the cross-track direction, given by
\begin{equation}
e_{path}=\left(-\sin\psi_{q}\left(\hat{p}_{n}-r_{n}\right)+\cos\psi_{q}\left(\hat{p}_{e}-r_{e}\right)\right)
\end{equation}
The parameters $\psi^{\infty}$ and $k_{path}$ determine the response of the guidance law to cross-track errors. The path unit vector $\boldsymbol{q}^{ned}$  and origin $\boldsymbol{r}=\left[\begin{array}{cc} r_{n} & r_{e}\end{array}\right]^{T}$ are determined by
\begin{equation}
\boldsymbol{r}=\boldsymbol{w}_{i-1}
\end{equation}
\begin{equation}
\boldsymbol{q}^{ned}=\frac{\boldsymbol{w}_{i+1}-\boldsymbol{w}_{i}}{\left\Vert \boldsymbol{w}_{i+1}-\boldsymbol{w}_{i}\right\Vert }
\end{equation}
where $\boldsymbol{w}_{i-1}$, $\boldsymbol{w}_{i}$, and $\boldsymbol{w}_{i+1}$ correspond to the previous, current, and subsequent way points, respectively. The heading of the desired path $\psi_q$ is calculated from the North and East components of the path unit vector as follows:
\begin{equation}
\psi_q = \textrm{atan2}\left(q_e,q_n\right)
\end{equation}
The low-level control law is comprised of a state-space system of integrator states,
\begin{equation}
\underset{\dot{\check{\boldsymbol{x}}}}{\underbrace{\left[\begin{array}{c}
\dot{\sigma}_{F}\\
\dot{\sigma}_{T}
\end{array}\right]}}=\underset{\check{\boldsymbol{f}}\left(\hat{\boldsymbol{x}},\hat{\boldsymbol{x}}^{*}\right)}{\underbrace{\left[\begin{array}{c}
V_{g}^{*}-\hat{V}_{g}\\
\psi^{*}-\hat{\psi}
\end{array}\right]}}
\end{equation}
and an output equation
\begin{equation}
\boldsymbol{u}=\left[\begin{array}{c}
F_{c}\\
T_{c}
\end{array}\right]=\underset{\boldsymbol{g}\left(\check{\boldsymbol{x}},\hat{\boldsymbol{x}},\hat{\boldsymbol{x}}^{*},\tilde{\boldsymbol{y}}\right)}{\underbrace{\left[\begin{array}{c}
P_{F}\left(V_{g}^{*}-\hat{V}_{g}\right)+I_{F}\sigma_{F}\\
D_{T}\left\{ P_{T}\left(\psi^{*}-\hat{\psi}\right)+I_{T}\sigma_{T}-\tilde{\omega}\right\} 
\end{array}\right]}}\label{eq:cntrl}
\end{equation}
where $P_F$ and $I_F$ are the proportional and integral control gains of the velocity controller.  Similarly, $P_T$, $I_T$, and $D_T$ are the proportional, integral, and derivative gains of the heading controller. Finally the mapping between truth states and navigation states is
\begin{equation}
\left[\begin{array}{c}
p_{n}\\
p_{e}\\
V_{g}\\
\psi
\end{array}\right]=\left[\begin{array}{cc}
I_{4\times4} & 0_{4\times3}\end{array}\right]\left[\begin{array}{c}
p_{n}\\
p_{e}\\
V_{g}\\
\psi\\
\omega\\
F_{dist}\\
T_{dist}
\end{array}\right]=\boldsymbol{m}\left(\boldsymbol{x}\right)
\end{equation}

\subsection{Linearized Model}\label{sec:lineared-vehicle-model}
The coefficient matrices in Eqs.~\eqref{eq:linmod4} through~\eqref{eq:linmod9} for the system under consideration are listed subsequently.  Linearization of Eq.~\eqref{eq:nonlin1} yields
\begin{equation}
F_{x}=\left[\begin{array}{ccccccc}
0 & 0 & \cos\bar{\psi} & -\bar{V}_{g}\sin\bar{\psi} & 0 & 0 & 0\\
0 & 0 & \sin\bar{\psi} & \bar{V}_{g}\cos\bar{\psi} & 0 & 0 & 0\\
0 & 0 & F_{V_{g}} & 0 & 0 & F_{u_{w}} & 0\\
0 & 0 & 0 & 0 & 1 & 0 & 0\\
0 & 0 & 0 & 0 & 0 & 0 & \frac{1}{J}\\
0 & 0 & -\frac{\bar{u}_{w}}{L_{u}} & 0 & 0 & -\frac{\bar{V}_{g}}{L_{u}} & 0\\
0 & 0 & 0 & 0 & 0 & 0 & -\frac{1}{\tau_{T}}
\end{array}\right]
\end{equation}
\begin{equation}
F_{u}=\left[\begin{array}{cc}
0 & 0\\
0 & 0\\
\frac{1}{m} & 0\\
0 & 0\\
0 & \frac{1}{J}\\
0 & 0\\
0 & 0
\end{array}\right]
\end{equation}
\begin{equation}
B=\left[\begin{array}{cc}
0 & 0\\
0 & 0\\
0 & 0\\
0 & 0\\
0 & 0\\
\sigma_{u}\sqrt{\frac{2\bar{V}_{g}}{L_{u}}} & 0\\
0 & 1
\end{array}\right]
\end{equation}
where
\begin{equation}
F_{V_{g}}=-\frac{\rho C_{D_{0}}S_{p}\left(\bar{V}_{g}-\bar{u}_{w}\right)}{m}
\end{equation}
\begin{equation}
F_{u_{w}}=\frac{\rho C_{D_{0}}S_{p}\left(\bar{V}_{g}-\bar{u}_{w}\right)}{m}
\end{equation}
The partial derivatives of the controller output Eq.~\eqref{eq:nonlin14} are
\begin{equation}
G_{\tilde{y}}=\left[\begin{array}{cc}
0 & 0\\
0 & -D_{T}
\end{array}\right]
\end{equation}
\begin{equation}
G_{\hat{x}}=\left[\begin{array}{cccc}
0 & 0 & -P_{F} & 0\\
0 & 0 & 0 & -D_{T}P_{T}
\end{array}\right]
\end{equation}
\begin{equation}
G_{\hat{x}^{*}}=\left[\begin{array}{cc}
P_{F} & 0\\
0 & D_{T}P_{T}
\end{array}\right]
\end{equation}
\begin{equation}
G_{\check{x}}=\left[\begin{array}{cc}
I_{F} & 0\\
0 & D_{T}I_{T}
\end{array}\right]
\end{equation}
The partial derivatives of the continuous measurement function in Eq.~\eqref{eq:nonlin2} are
\begin{equation}
C_{x}=\left[\begin{array}{ccccccc}
0 & 0 & -\frac{\rho C_{D_{0}}S_{p}\left(\bar{V}_{g}-\bar{u}_{w}\right)}{m} & 0 & 0 & \frac{\rho C_{D_{0}}S_{p}\left(\bar{V}_{g}-\bar{u}_{w}\right)}{m} & 0\\
0 & 0 & 0 & 0 & 1 & 0 & 0
\end{array}\right]
\end{equation}
\begin{equation}
C_{u}=\left[\begin{array}{cc}
\frac{1}{m} & 0\\
0 & 0
\end{array}\right]
\end{equation}
The partial derivative of the guidance law of Eq.~\eqref{eq:nonlin12} is
\begin{equation}
N_{\hat{x}}=\left[\begin{array}{cccc}
0 & 0 & 0 & 0\\
N_{21}/N_D & N_{22}/N_D & 0 & 0
\end{array}\right]\label{eq:lin_nx}
\end{equation}
where
\begin{equation}
N_{21}=2\psi^{\infty}k_{path}\sin\psi_{q}
\end{equation}
\begin{equation}
N_{22}= -2\psi^{\infty}k_{path}\cos\psi_{q}
\end{equation}
\begin{equation}
N_D = \pi+\pi k_{path}^{2}d^2
\end{equation}
\begin{equation}
    d=-\sin\psi_{q}\left[\hat{\bar{p}}_{n}-r_{n}\right]+\cos\psi_{q}\left[\hat{\bar{p}}_{e}-r_{e}\right]
\end{equation}
\begin{equation}
    N_{\tilde{y}}=0_{\hat{n} \times n_{\tilde{y}}}
\end{equation}
The partial derivatives of the navigation state propagation function in Eq.~\eqref{eq:nonlin5} are
\begin{equation}
\hat{F}_{\hat{x}}=\left[\begin{array}{cccc}
0 & 0 & \cos\hat{\psi} & -\hat{V}_{g}\sin\hat{\psi}\\
0 & 0 & \sin\hat{\psi} & \hat{V}_{g}\cos\hat{\psi}\\
0 & 0 & 0 & 0\\
0 & 0 & 0 & 0
\end{array}\right]
\end{equation}
\begin{equation}
\hat{F}_{\tilde{y}}=\left[\begin{array}{cc}
0 & 0\\
0 & 0\\
1 & 0\\
0 & 1
\end{array}\right]
\end{equation}
The partials of the controller propagation function in Eq.~\eqref{eq:nonlin13} are 
\begin{equation}
\check{F}_{\hat{x}}=\left[\begin{array}{cccc}
0 & 0 & -1 & 0\\
0 & 0 & 0 & -1
\end{array}\right]
\end{equation}
\begin{equation}
\check{F}_{\hat{x}^{*}}=\left[\begin{array}{cc}
1 & 0\\
0 & 1
\end{array}\right]
\end{equation}
The partials of the truth and navigation measurement models in Eqs.~\eqref{eq:nonlin4} and~\eqref{eq:nonlin7} are
\begin{equation}
H_{x}=\left[\begin{array}{ccccccc}
1 & 0 & 0 & 0 & 0 & 0 & 0\\
0 & 1 & 0 & 0 & 0 & 0 & 0\\
0 & 0 & 1 & 0 & 0 & 0 & 0
\end{array}\right]
\end{equation}
\begin{equation}
\hat{H}_{\hat{x}}=\left[\begin{array}{cccc}
1 & 0 & 0 & 0\\
0 & 1 & 0 & 0\\
0 & 0 & 1 & 0
\end{array}\right]
\end{equation}
Finally, the partial of the mapping between truth and navigation states in Eq.~\eqref{eq:nonlin11} is
\begin{equation}
M_{x}=\left[\begin{array}{cc}
I_{4\times4} & 0_{4\times3}\\
0_{3\times4} & 0_{3\times3}
\end{array}\right]
\end{equation}


\section{Path Planning\label{subsec:Path-Planning}}
This section of the paper defines the approach for path planning in a \revision{static,} uncertain obstacle field.  A method is developed for computing the probability of collision from the truth state dispersion covariance and a probabilistic description of the obstacle field.  This method forms the basis of the \textit{CollisionFree} primitive for the Rapidly-exploring Random Tree path planner.
\subsection{Probability of Collision\label{subsec:probability-of-collision}}
\revision{The chance constraint considered in this research is the probability of collision between the vehicle and a static obstacle with known size and shape and uncertain location.  Many approaches exist for determining collision probabilities, each accommodating different sizes and shapes of the vehicle and/or obstacle, as well as uncertainty in the state of the vehicle and/or the obstacle~\cite{bry_rapidly-exploring_2011,park_efficient_2017,park_fast_2016,inaba_safe_2016,van_den_berg_lqg-mp_2011,wu_probabilistically_2020,okamoto_optimal_2019,zhu_chance-constrained_2019,du_toit_probabilistic_2011,summers_distributionally_2018,castillo-lopez_real-time_2020,blackmore_chance-constrained_2011}.  A common approach is to model the vehicle as a point and the obstacle as a polyhedron.  In this scenario, the path planner ensures that the probability of the vehicle position being inside the polyhedron is below a user-specified threshold, often by converting the collision chance constraint to a series of deterministic constraints on the vehicle mean, one for each side of the polyhedron~\cite{blackmore_chance-constrained_2011,du_toit_probabilistic_2011}.  Alternatively, the probability of collision may be computed directly from the Gaussian PDF, as is the case in this work.  The polytope assumed in this manuscript is a square bounding box with known size; the box's position follows a Gaussian distribution.  To aid in understanding this paradigm of collision probability, consider two \emph{scalar}, \emph{independent} Gaussian random variables, representing the position of two objects,}
\begin{equation}
x\sim N\left(m_{x},P_{x}\right)
\end{equation}
\begin{equation}
y\sim N\left(m_{y},P_{y}\right)
\end{equation}

Combining the random variables as a vector yields
\begin{equation}
\boldsymbol{z}=\left[\begin{array}{cc}
x & y\end{array}\right]^{T}\in\mathbb{R}^{2}
\end{equation}
where the joint PDF is defined as
\begin{equation}
f_{\boldsymbol{z}}\left(\boldsymbol{\zeta}\right)=\frac{1}{N_{z}}\exp\left\{ -\frac{1}{2}\left[\boldsymbol{\zeta}-\boldsymbol{m}_{z}\right]^{T}P_{zz}^{-1}\left[\boldsymbol{\zeta}-\boldsymbol{m}_{z}\right]\right\} 
\end{equation}
\begin{equation}
N_{z}=2\pi\left|P_{zz}\right|^{1/2}
\end{equation}
Since the two random variables are independent, the mean and covariance of the joint distribution are simply the augmentation of the individual parameters
\begin{equation}
\boldsymbol{m}_{z}=\left[\begin{array}{cc}
m_{x} & m_{y}\end{array}\right]^{T}
\end{equation}
\begin{equation}
P_{zz}=\left[\begin{array}{cc}
P_{x} & 0\\
0 & P_{y}
\end{array}\right]
\end{equation}

The relative position of the two objects is computed as a linear operation on the random vector $\boldsymbol{z}$
\begin{equation}
d=A\boldsymbol{z}
\end{equation}
where 
\begin{equation}
A=\left[\begin{array}{cc}
1 & -1\end{array}\right]
\end{equation}
This corresponds to a new Gaussian random variable with the following mean, covariance, and probability density function~\cite{maybeck_stochastic_1994}
\begin{equation}
m_{d}=A\boldsymbol{m}_{z}
\end{equation}
\begin{equation}
P_{d}=AP_{zz}A^{T}
\end{equation}
\begin{equation}
f_{d}\left(\delta\right)=\frac{1}{\sqrt{2\pi P_{d}}}\exp\left\{ -\frac{1}{2P_{d}}\left(\delta-m_{d}\right)^{2}\right\} 
\end{equation}

Finally, the probability that the two objects are closer than a specified threshold $l$ is the integration of $f_{d}$ over $\delta=\pm l$, i.e.,
\begin{eqnarray}
P\left(l\right) \equiv \intop_{-l}^{l}f_{d}\left(\rho\right)d\rho
\end{eqnarray}

Extension of this collision probability paradigm to two dimensions with $N$ obstacles is as follows. The combined system is expressed as a random vector of the vehicle position $p$ and the $N$ obstacle positions,
\begin{equation}
\boldsymbol{z}=\left[\begin{array}{c}
\boldsymbol{p}\\
\boldsymbol{p}_{1}\\
\boldsymbol{p}_{2}\\
\vdots\\
\boldsymbol{p}_{N}
\end{array}\right]\in\mathbb{R}^{2\left(N+1\right)}
\end{equation}
with probability density function, mean, and covariance defined as
\begin{equation}
f_{\boldsymbol{z}}\left(\boldsymbol{\zeta}\right)=\frac{1}{N_{z}}\exp\left\{ -\frac{1}{2}\left[\boldsymbol{\zeta}-\boldsymbol{m}_{z}\right]^{T}P_{zz}^{-1}\left[\boldsymbol{\zeta}-\boldsymbol{m}_{z}\right]\right\} 
\end{equation}
\begin{equation}
N_{z}=\left(2\pi\right)^{\left(N+1\right)}\left|P_{zz}\right|^{1/2}
\end{equation}
\begin{equation}
\boldsymbol{m}_{z}=\left[\begin{array}{c}
\boldsymbol{m}_{uav}\\
\boldsymbol{m}_{1}\\
\boldsymbol{m}_{2}\\
\vdots\\
\boldsymbol{m}_{N}
\end{array}\right]\label{eq:jointmean}
\end{equation}
\begin{equation}
P_{zz}=\textrm{diag}\left(\left[\begin{array}{ccccc}
P_{uav} & P_{1} & P_{2} & \cdots & P_{N}\end{array}\right]\right)\label{eq:jointcov}
\end{equation}

The relative position of the vehicle with respect to obstacle $i$ is computed as a linear transformation of the random vector $\boldsymbol{z}$ , i.e.,
\begin{equation}
\boldsymbol{d}=A\boldsymbol{z}
\end{equation}
where 
\begin{equation}
A=\left[\begin{array}{cc}
I_{2\times2} & A_{i}\end{array}\right]
\end{equation}
\begin{equation}
A_{i}=\left[\begin{array}{ccc}
0_{2\times2\left(i-1\right)} & -I_{2\times2} & 0_{2\times2\left(N-i\right)}\end{array}\right]
\end{equation}
This calculation yields a new Gaussian random vector with the following mean, covariance, and distribution:
\begin{equation}
\boldsymbol{m}_{d}=A\boldsymbol{m}_{z}
\end{equation}
\begin{equation}
P_{d}=AP_{zz}A^{T}
\end{equation}
\begin{equation}
f_{\boldsymbol{d}}\left(\boldsymbol{\delta}\right)=N_d\exp\left\{ -\frac{1}{2}\left[\boldsymbol{\delta}-\boldsymbol{m}_{d}\right]^{T}P_{dd}^{-1}\left[\boldsymbol{\delta}-\boldsymbol{m}_{d}\right]\right\} 
\end{equation}
\begin{equation}
N_{d}=\frac{1}{2\pi\left|P_{dd}\right|^{1/2}}
\end{equation}
Finally, the probability that the vehicle is closer than the vector threshold $\boldsymbol{l}$ is the integration of $f_{\boldsymbol{d}}$ over $\boldsymbol{\delta}=\pm\boldsymbol{l}$, i.e.,
\begin{eqnarray}\label{eq:p_collision}
P\left(\boldsymbol{l}\right) \equiv \intop_{-l_{1}}^{l_{1}}\intop_{-l_{2}}^{l_{2}}f_{\boldsymbol{d}}\left(\lambda_{1},\lambda_{2}\right)d\lambda_{1}d\lambda_{2}
\end{eqnarray}


\subsection{RRT with Probability of Collision Constraint\label{subsec:RRT+P_c}}
\revision{Many options exist for planning the path of a vehicle through an obstacle field. An excellent review of path planning, with a focus on sampling-based planners is provided in~\cite{karaman_sampling-based_2011}, where rapidly exploring random trees (RRTs) are cited as particularly suitable for vehicles with differential constraints, nonlinear dynamics, and non-holonomic constraints.  As its name suggests, the RRT allows for a rapid search through high-dimensional spaces. In its most simple form, this algorithm iteratively builds a tree while  randomly searching through the environment. In each iteration, a point is randomly selected and the nearest point on the tree is used to attempt to connect to the newly selected point. When an obstacle precludes the actual connecting of the tree to the new point, a point along the new path prior to the obstacle intersection is added to the tree.  The RRT exhibits useful properties, two of which are probabilistic completeness and an exponentially decaying probability of failure~\cite{lavalle_randomized_2001}.}

\revision{One of the aspects that makes RRT such a versatile planner is that it is expressed in terms of a number of primitive procedures. Karaman and Frazzoli~\cite{karaman_sampling-based_2011} present a concise overview of many of these procedures for describing RRT. \textit{Sample} selects a new point, \textit{Nearest} finds vertices in the tree close to a particular point, and \textit{Steer} attempts to connect different points.  Many of the advancements in RRT and its almost surely asymptotically optimal counterpart RRT* have focused on improvements to these primitives~\cite{gammell_informed_2014,nasir_rrt-smart_2013,moon_kinodynamic_2015,beard_randy_small_2012,yang_spline-based_2014,noreen_optimal_2016,arslan_sampling-based_2017}.  Common to all RRT planners is the \textit{CollisionFree} primitive, which verifies the absence of obstacles along a candidate path.  The RRT algorithm implemented in this research is a combination of the RRT algorithms discussed in~\cite{karaman_sampling-based_2011,beard_randy_small_2012}, and is documented in Algorithm~\ref{alg:Rapidly-Exploring-Random}.}\newline

\begin{algorithm2e}[H]
\SetInd{1em}{1em}
\DontPrintSemicolon
\caption{Rapidly Exploring Random Trees\label{alg:Rapidly-Exploring-Random}}
\SetAlgoLined
\KwResult{Selected waypoint path $\mathcal{W}$} 
	Initialize directed graph $G=(V,E)$ where $V=\left\{ \begin{array}{cc} \boldsymbol{x}_{s} & \boldsymbol{x}_{f}\end{array}\right\}$, $E=\emptyset$\;
	Initialize obstacle map $\mathcal{O}=\left\{ \boldsymbol{m_{i}},P_{i}\right\}$\;
	\While{$i<N$}{
	 $\boldsymbol{x}_{rand}\leftarrow\textit{Sample}\left(i\right)$\;
	 $\boldsymbol{x}_{nearest}\leftarrow\textit{Nearest}\left(p,V\right)$\;
	 $\boldsymbol{x}_{new}\leftarrow\textit{Steer}\left(\boldsymbol{x}_{nearest},\boldsymbol{x}_{rand},\lambda\right)$\;
	 $\boldsymbol{x}\left(t\right),\Delta t\leftarrow\textit{RunSim}\left(\boldsymbol{x}_{nearest},\boldsymbol{x}_{new}\right)$\;
	 \If{$\textit{CollisionFree}\left(\boldsymbol{x}\left(t\right),\mathcal{O}\right)$}{
	  $V\leftarrow V\cup\boldsymbol{x}_{new}$\;
	  $E\leftarrow E\cup \left\{ \left(\boldsymbol{x}_{nearest},\boldsymbol{x}_{new}\right),\Delta t\right\}$\;
	  }
	 \If{$\left\Vert \boldsymbol{x}_{new}-\boldsymbol{x}_{f}\right\Vert \leq\lambda$}{
	  $\boldsymbol{x}\left(t\right),\Delta t\leftarrow\textit{RunSim}\left(\boldsymbol{x}_{new},\boldsymbol{x}_{f}\right)$\;
	  \If{$\textit{CollisionFree}\left(\boldsymbol{x}\left(t\right),\mathcal{O}\right)$}{
	   $V\leftarrow V\cup\boldsymbol{x}_{f}$\;
	   $E\leftarrow E\cup \left\{ \left(\boldsymbol{x}_{new},\boldsymbol{x}_{f}\right),\Delta t\right\}$\;
	  }
	 }
 }
$\mathcal{W}\leftarrow \textit{SelectPath}\left(G\right)$\;
\Return{$\mathcal{W}$}
\end{algorithm2e}

Some discussion regarding the primitives in Algorithm~\ref{alg:Rapidly-Exploring-Random} is warranted. The primitive $\textit{Sample}$ of line 4 randomly samples the UAV position for a user-specified range of north/east values, i.e.,
\begin{equation}
    \boldsymbol{x}_{rand}\sim U\left(\boldsymbol{x}_{min},\boldsymbol{x}_{max}\right)
\end{equation}
The primitive $\textit{Nearest}$ finds the vertex in $V$ whose position is closest to $\boldsymbol{x}_{rand}$, in an $L_{2}$ norm sense,
\begin{equation}
    \argmin_{\boldsymbol{x}_{nearest}} \norm{\boldsymbol{x}_{nearest} - \boldsymbol{x}_{rand}}
\end{equation}
The primitive $\textit{Steer}$ returns a position beginning at $\boldsymbol{x}_{nearest}$ in the direction of $\boldsymbol{x}_{rand}$ with a magnitude of $\lambda$, given by
\begin{equation}
    \boldsymbol{x}_{new} = \boldsymbol{x}_{nearest} + \frac{\boldsymbol{x}_{rand}}{\norm{\boldsymbol{x}_{rand}}}\lambda
\end{equation}
The primitive $\textit{RunSim}$ calculates the vehicle trajectory $\boldsymbol{x}(t)$  by propagating the nonlinear system in Section \ref{sec:NonlinearVehicleModel} from $\boldsymbol{x}_{nearest}$ to $\boldsymbol{x}_{new}$ with all noise terms set to zero, which defines the nominal reference trajectory $\bar{\boldsymbol{x}}$ to be used in the LinCov simulation. The primitive $\textit{CollisionFree}$ propagates the augmented system of~\eqref{eq:augmented-diff-eq} and~\eqref{eq:augmented-update}, as specified by the coefficient matrices of in Section \ref{sec:lineared-vehicle-model}, along the nominal reference trajectory $\bar{\boldsymbol{x}}$ producing the covariance of the vehicle state dispersions $D_{true}$. The nominal vehicle position and dispersion covariance are computed as
\revision{\begin{equation}
    \label{eq:uav-mean}
    \boldsymbol{m}_{uav}(t) = A_{uav}\bar{\boldsymbol{x}}(t)
\end{equation}
\begin{equation}
    \label{eq:uav-cov}
    P_{uav} = A_{uav}D_{true}A_{uav}^T
\end{equation}
where
\begin{equation}
    \label{eq:uav-A}
    A_{uav} = \left[\begin{array}{cc} I_{2\times2} & 0_{2\times5}\end{array}\right]
\end{equation}}
\revision{Combining Eqs.~\eqref{eq:uav-mean}-~\eqref{eq:uav-A}  with each obstacle mean $m_i$ and covariance $P_i$ in the obstacle map $\mathcal{O}$} yields the system described in Eqs.~\eqref{eq:jointmean} and~\eqref{eq:jointcov}. The probability of collision is computed at every time step along the nominal reference trajectory as per equation~\eqref{eq:p_collision}. If the probability of collision is below a user-specified threshold, the path is considered free of obstacles. Finally, the primitive $\textit{SelectPath}$ searches the graph for the shortest distance path from the start vertex $\boldsymbol{x}_s$ to the final vertex $\boldsymbol{x}_f$.

\section{Simulation Results\label{sec:Simulation-Results}}
This section documents the results of the simulation. Section \ref{subsec:LinCov-Validation} validates the LinCov model developed in Section \ref{subsec:Vehicle-Model} by comparing the trajectory dispersions and estimation errors obtained from Monte Carlo analysis with the covariance computed using the LinCov simulation. Section \ref{subsec:Path-Planning-Demonstration} demonstrates the augmentation of the RRT path planning algorithm with the uncertain obstacles and UAV position dispersions covariance discussed in Section \ref{subsec:Path-Planning}.

Tables \ref{tab:vehicle-params}--\ref{tab:ctrl-params} list the many parameters used in the simulation.  The vehicle parameters in Table \ref{tab:vehicle-params} are based on the Aerosonde UAV~\cite{beard_randy_small_2012}.  The gust model parameters in Table \ref{tab:dist-params} are based on the Dryden parameters for low altitude and light turbulence~\cite{langelaan_wind_2011}.  The disturbance torque parameters in Table \ref{tab:dist-params} are not based on physical modeling; they are sized to contribute significantly to the vehicle dispersion and exercise the overall framework.  Table \ref{tab:sens-params} defines the noise of the inertial measurement unit and position/velocity sensors.  All parameters are typical of GPS-aided inertial navigation systems, with the exception of the gyro noise, which, similar to the disturbance torques, is sized to exercise the developed LinCov and path planning framework.  Finally, Table \ref{tab:ctrl-params} lists the parameters used inside the guidance and control laws.

\begin{table}[hbt!]
\caption{\label{tab:vehicle-params} Vehicle Parameters}
\centering
\begin{tabular}{lcccccc}
\hline
Parameter & Symbol & Value & Units\\
\hline
Nominal velocity & $\bar{V}_{g}$ & 35.0 & $m/s$\\
Air density & $\rho$ & 1.2682  & $kg/m^{3}$\\
Drag coefficient & $C_{D_{0}}$ & 0.03 & NA\\
Planform Area & $S_{p}$ & 0.55 & $m^{2}$\\
Mass & $m$ & 25.0 & $kg$\\
Inertia & $J$ & 1.759  & $kg\cdot m^{2}$\\
\hline
\end{tabular}
\end{table}

\begin{table}[hbt!]
\caption{\label{tab:dist-params} Disturbance Parameters}
\centering
\begin{tabular}{lcccccc}
\hline
Parameter & Symbol & Value & Units\\
\hline
Gust stdev & $\sigma_{u}$ & 1.06 & $m/s$\\
Gust correlation distance & $L_{u}$ & 200 & $m$\\
Disturbance torque stdev & $\sigma_{T}$ & 0.0033 & $N\cdot m$\\
Disturbance torque correlation time & $\tau_{T}$ & 2.0 & $s$\\
\hline
\end{tabular}
\end{table}

\begin{table}[hbt!]
\caption{\label{tab:sens-params} Sensor Parameters}
\centering
\begin{tabular}{lcccccc}
\hline
Parameter & Symbol & Value & Units\\
\hline
Velocity random walk & $\sqrt{S_{a}}$ & 0.02 & $m/s/\sqrt{hr}$\\
Angular random walk & $\sqrt{S_{\omega}}$ & 16.7 & $deg/\sqrt{hr}$\\
Position measurement noise & $\sqrt{R_{p}}$ & 1.0 & $m$\\
Velocity measurement noise & $\sqrt{R_{v}}$ & 0.033 & $m/s$\\
\hline
\end{tabular}
\end{table}

\begin{table}[hbt!]
\caption{\label{tab:ctrl-params} Controller Parameters}
\centering
\begin{tabular}{lcccccc}
\hline
Parameter & Symbol & Value & Units\\
\hline
Velocity proportional gain & $P_{F}$ & 80.0 & $kg/s$\\
Velocity integral gain & $I_{F}$ & 50.0 & $kg/s^{2}$\\
Attitude proportional gain & $P_{T}$ & 6.38 & $1/s$\\
Attitude integral gain & $I_{T}$ & 10.0 & $1/s^{2}$\\
Attitude derivative gain & $D_{T}$ & 111.0 & $kg\cdot m^{2}/s$\\
Path control angle & $\psi_{\infty}$ & $\pi/2$ & $rad$\\
Path control gain & $k_{path}$ & 0.05 & $1/m$\\
\hline
\end{tabular}
\end{table}

\subsection{LinCov Validation\label{subsec:LinCov-Validation}}
\begin{figure}[hbt!]
\centering\includegraphics[width=3.25in]{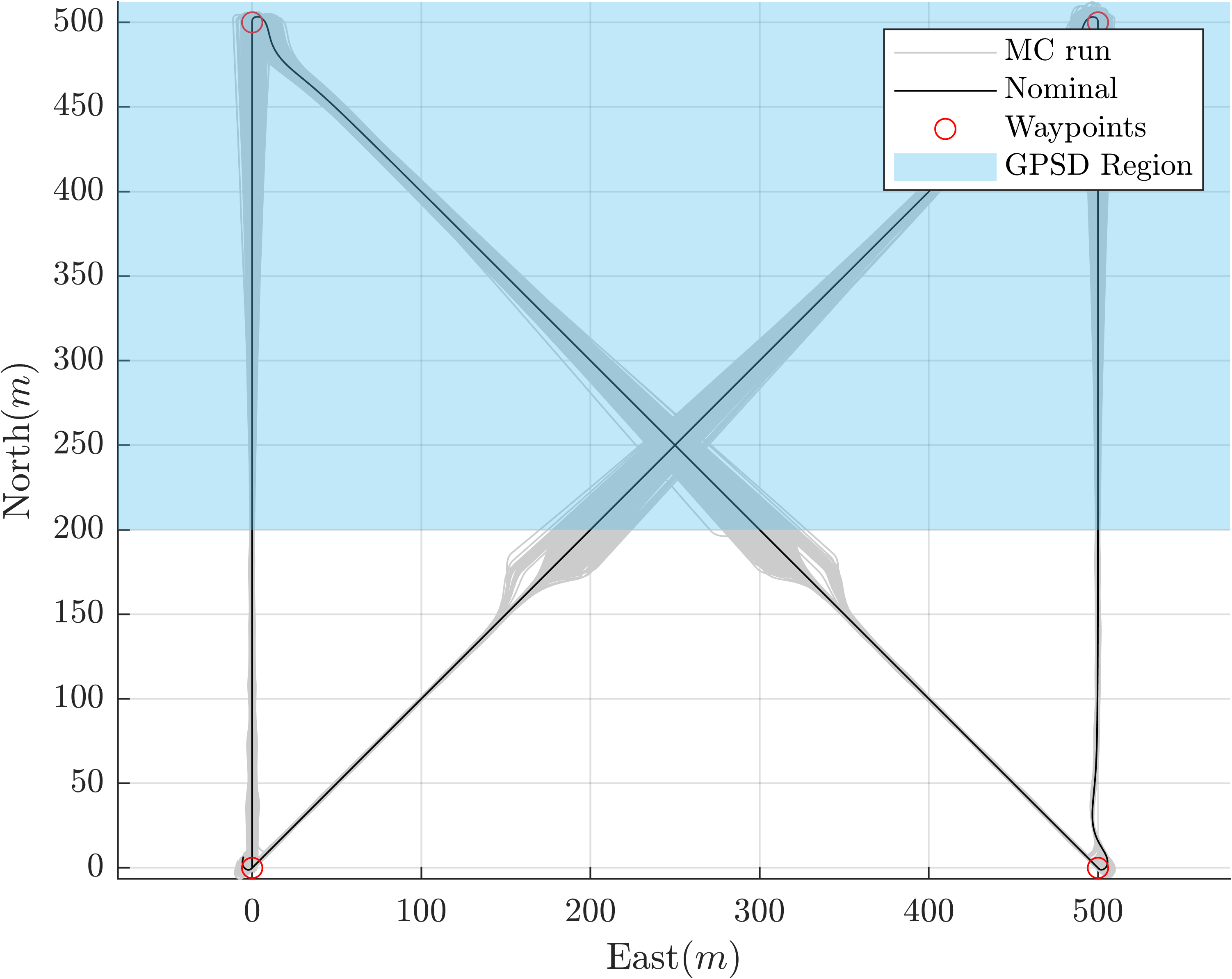}
\caption{UAV trajectory with shaded GPS-denied region\label{fig:uav_traj}}
\end{figure}

This section of the paper serves to validate the Linear Covariance model \revision{defined in Section~\ref{subsec:Linear-Modeling} with the problem-specific coefficient matrices defined in Section~\ref{sec:lineared-vehicle-model}}.  The method for validation is comparison of the truth state dispersion covariance and the estimation error covariance to the ensemble statistics of a 500-run Monte Carlo simulation.  \revision{The Monte Carlo simulation is executed using the \emph{nonlinear} models in Section \ref{sec:NonlinearVehicleModel}, with random initial conditions, process noise, measurement noise, etc.  In contrast, the covariance values are computed using the Linear Covariance framework derived in Sections \ref{subsec:Linear-Modeling} and \ref{subsec:Performance-Evaluation}, with coefficient matrices defined in Section \ref{sec:lineared-vehicle-model}}.  Figure \ref{fig:uav_traj} illustrates the scenario used in the validation, which consists of the UAV traveling along a set of waypoints, transitioning in and out of the blue GPS-denied region, where position and ground speed measurements are unavailable.  The black lines show the nominal path of the vehicle, with each individual Monte Carlo run shown in grey.

\begin{figure}[hbt!]
\centering
\includegraphics[width=3.25in]{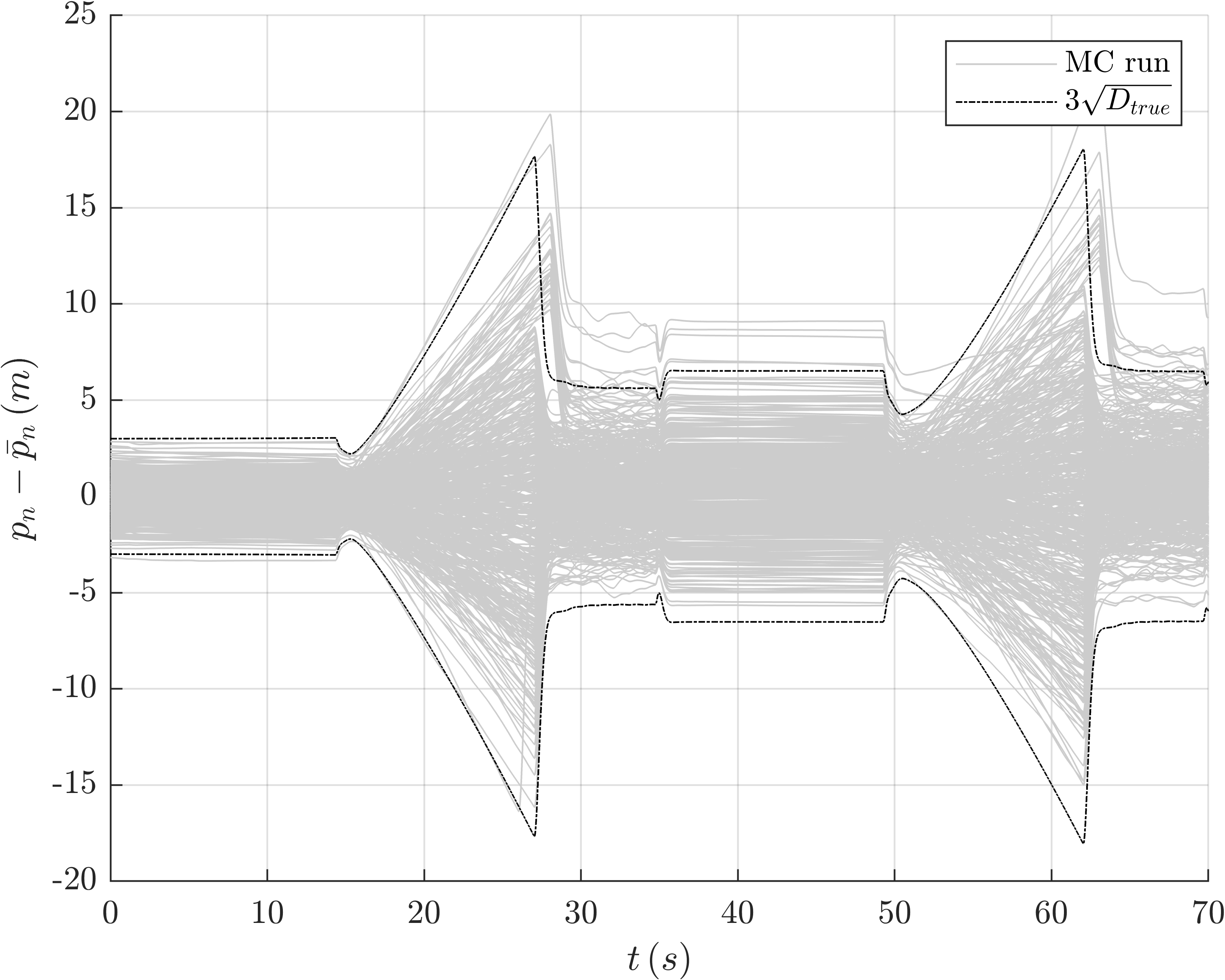}
\caption{North position dispersions\label{fig:Position-dispersion-north}}
\end{figure}

\begin{figure}[hbt!]
\centering
\includegraphics[width=3.25in]{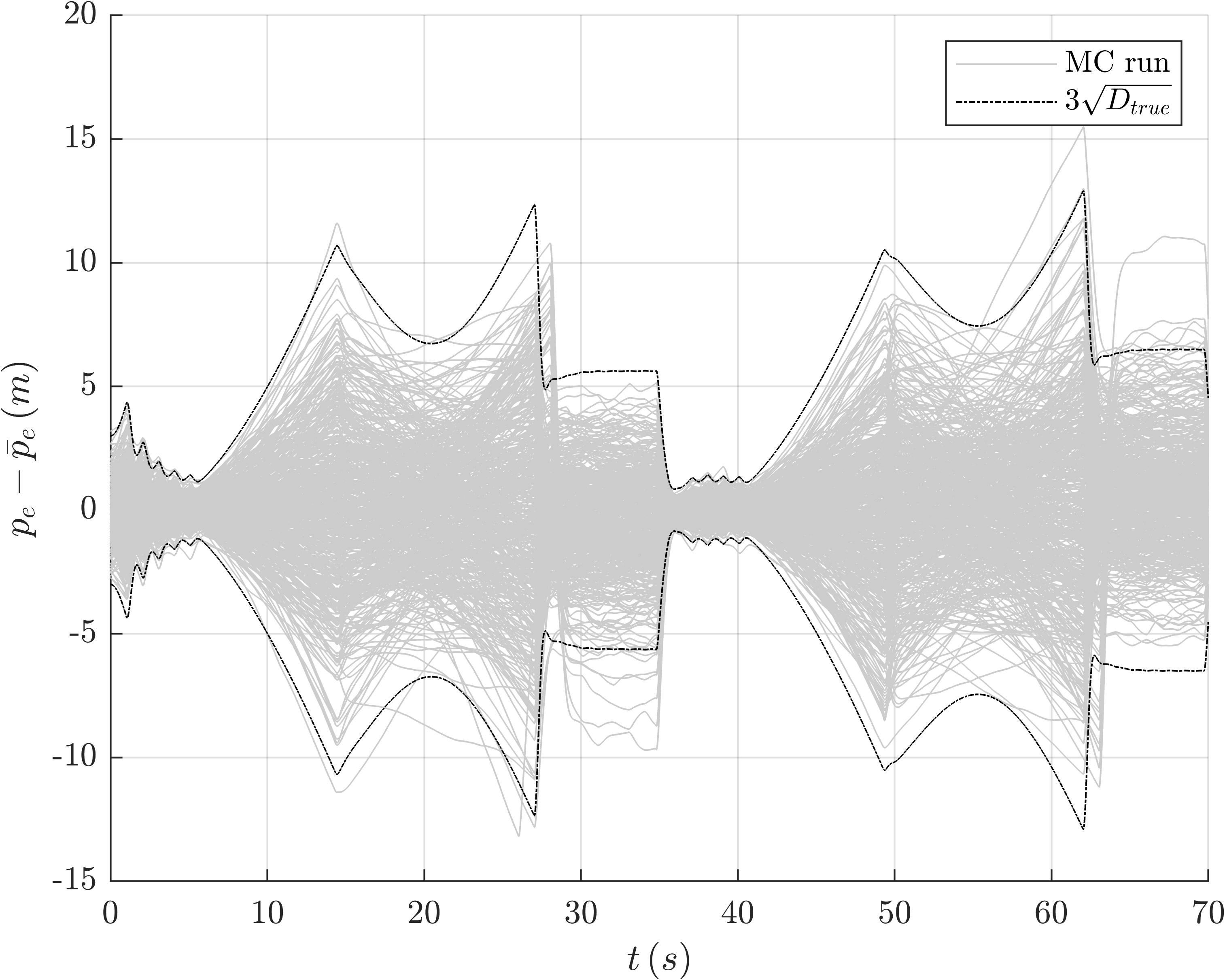}
\caption{East position dispersions\label{fig:Position-dispersion-east}}
\end{figure}

\begin{figure}[hbt!]
\centering
\includegraphics[width=3.0in]{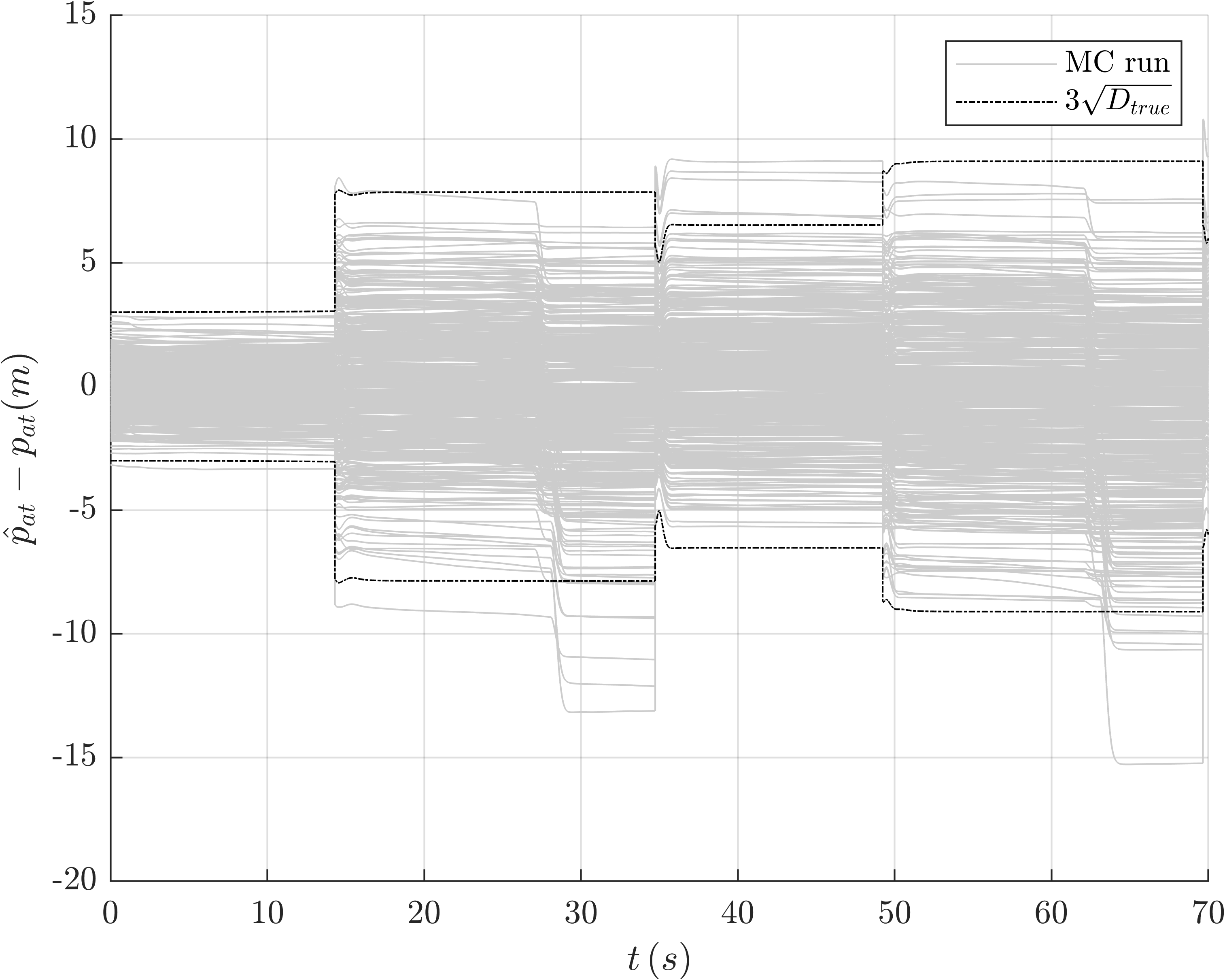}
\caption{Along track position dispersions\label{fig:Position-dispersion-along-track}}
\end{figure}

\begin{figure}[hbt!]
\centering
\includegraphics[width=3.0in]{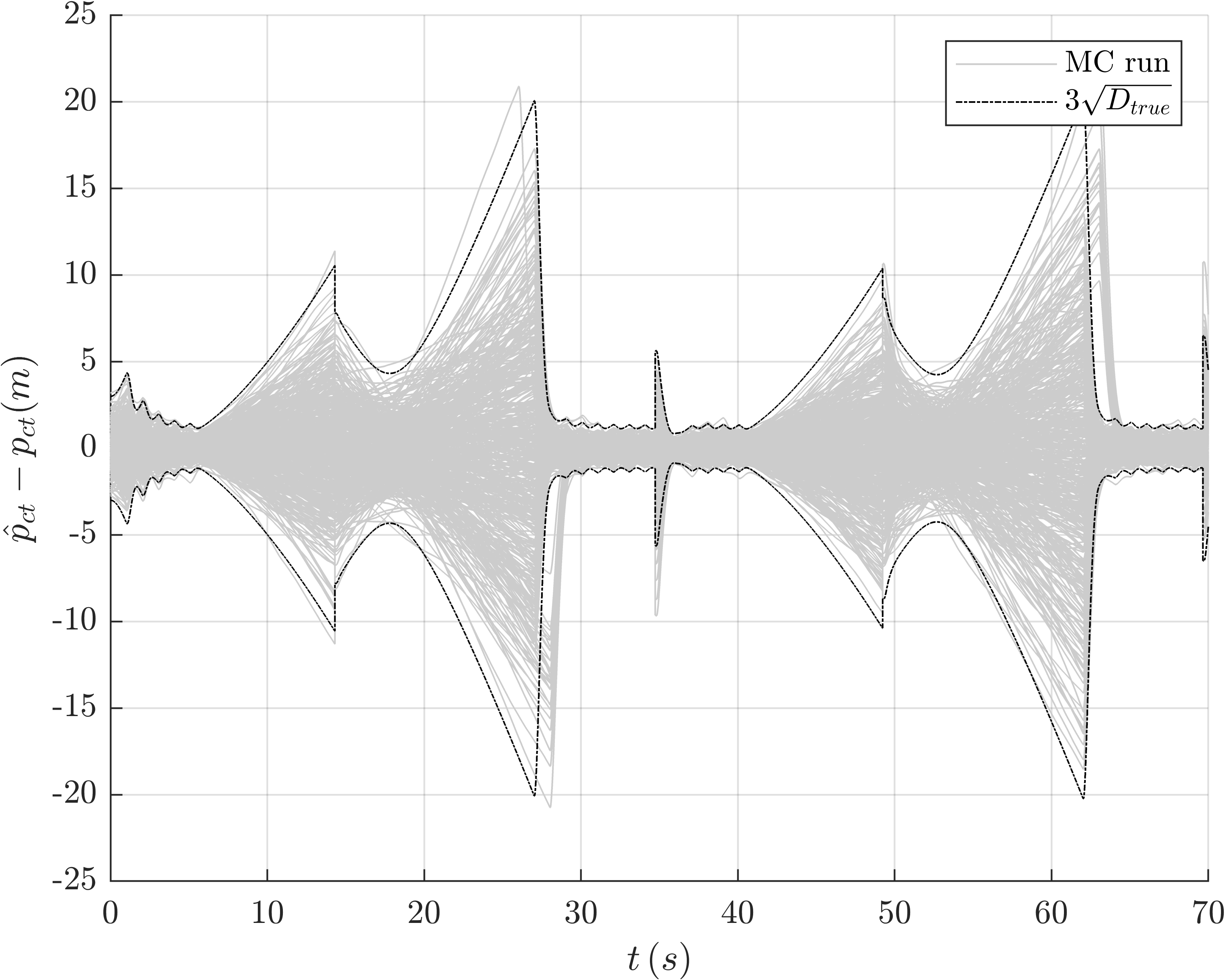}
\caption{Cross-track position dispersions\label{fig:Position-dispersion-cross-track}}
\end{figure}

\begin{figure}[hbt!]
\centering
\includegraphics[width=3.0in]{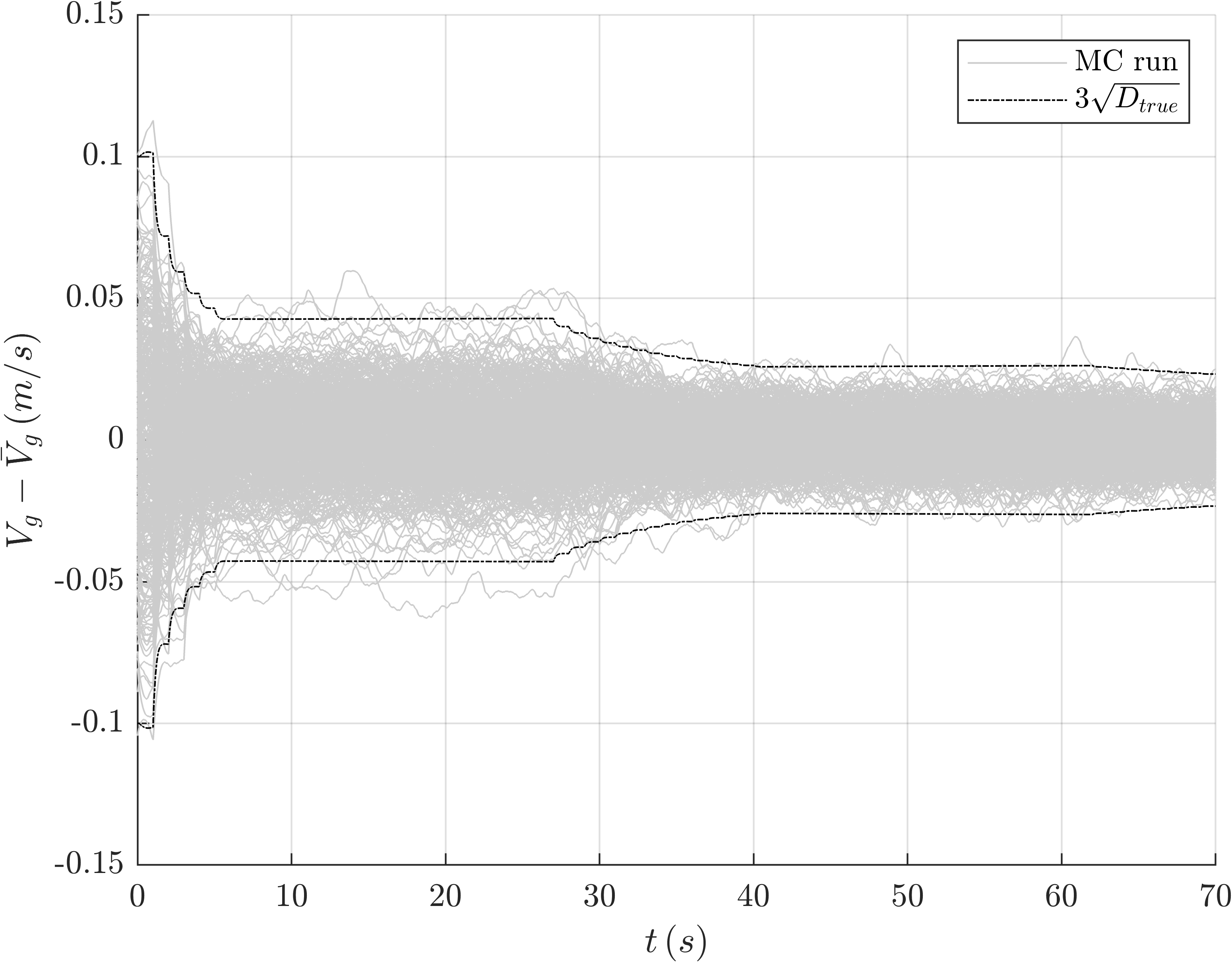}
\caption{Ground speed dispersions\label{fig:velocity_dispersion}}
\end{figure}

\begin{figure}[hbt!]
\centering
\includegraphics[width=3.0in]{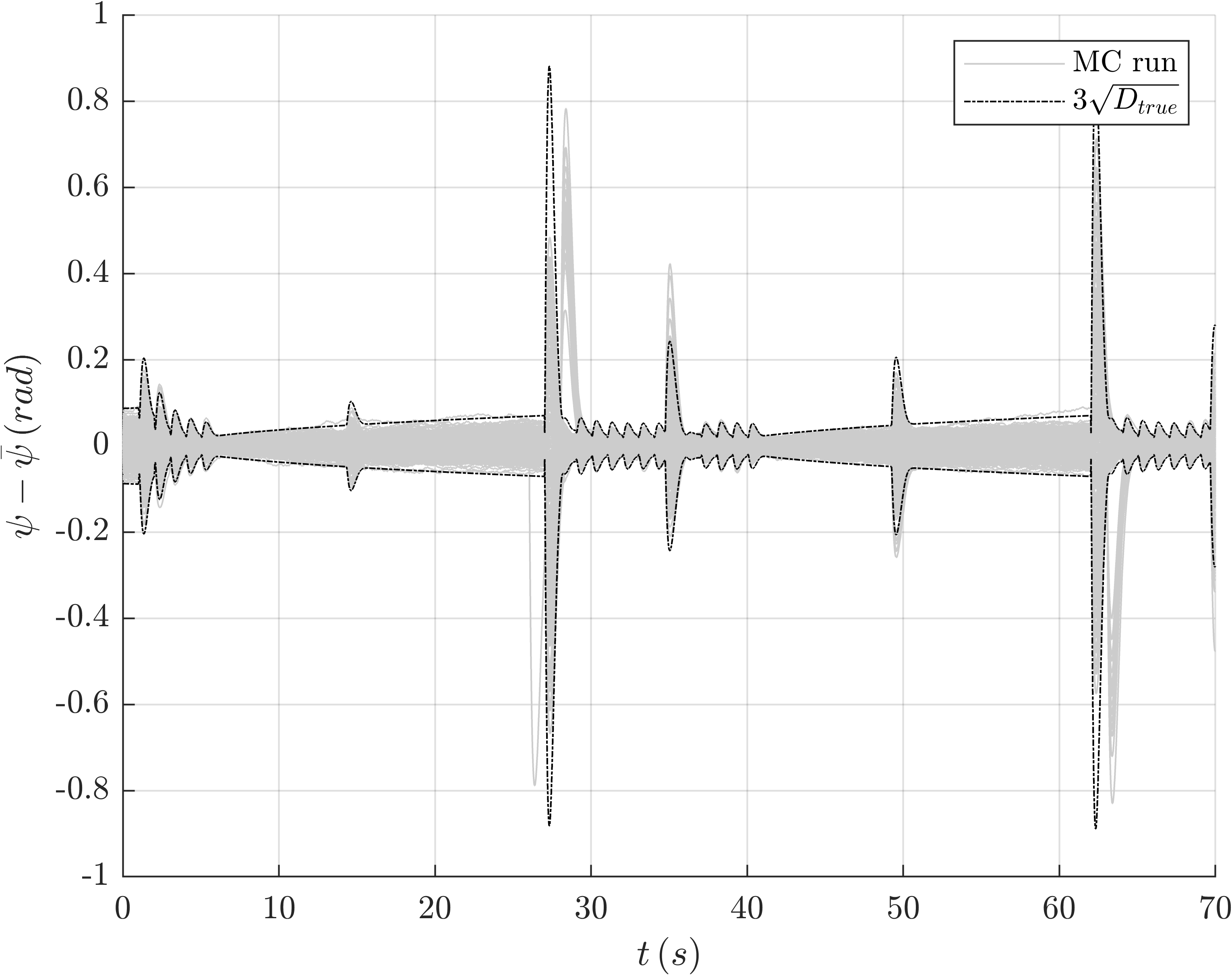}
\caption{Heading dispersions\label{fig:heading_dispersion}}
\end{figure}

\begin{figure}[hbt!]
\centering
\includegraphics[width=3.0in]{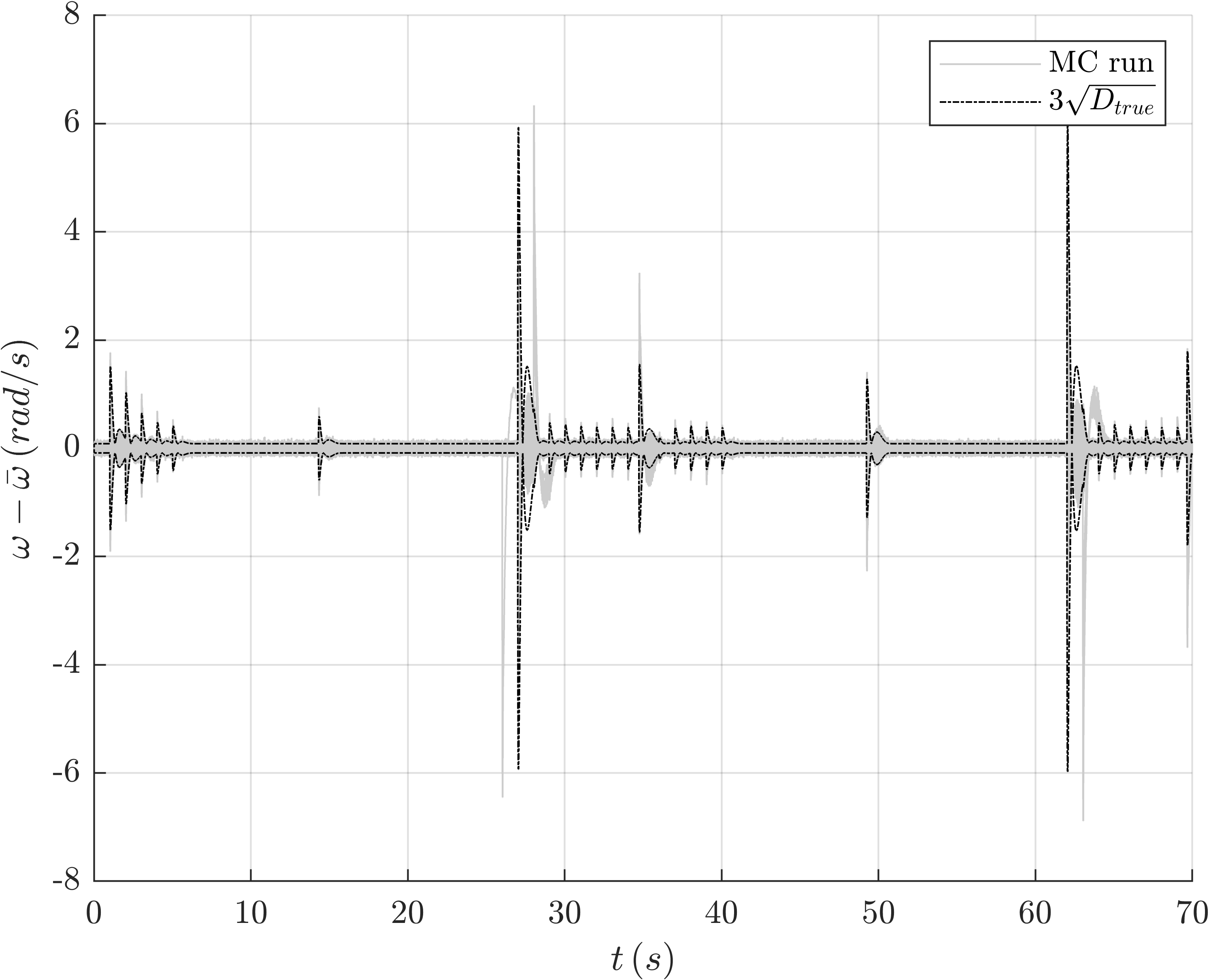}
\caption{Angular rate dispersions\label{fig:angular_rate_dispersion}}
\end{figure}

\begin{figure}[hbt!]
\centering
\includegraphics[width=3.0in]{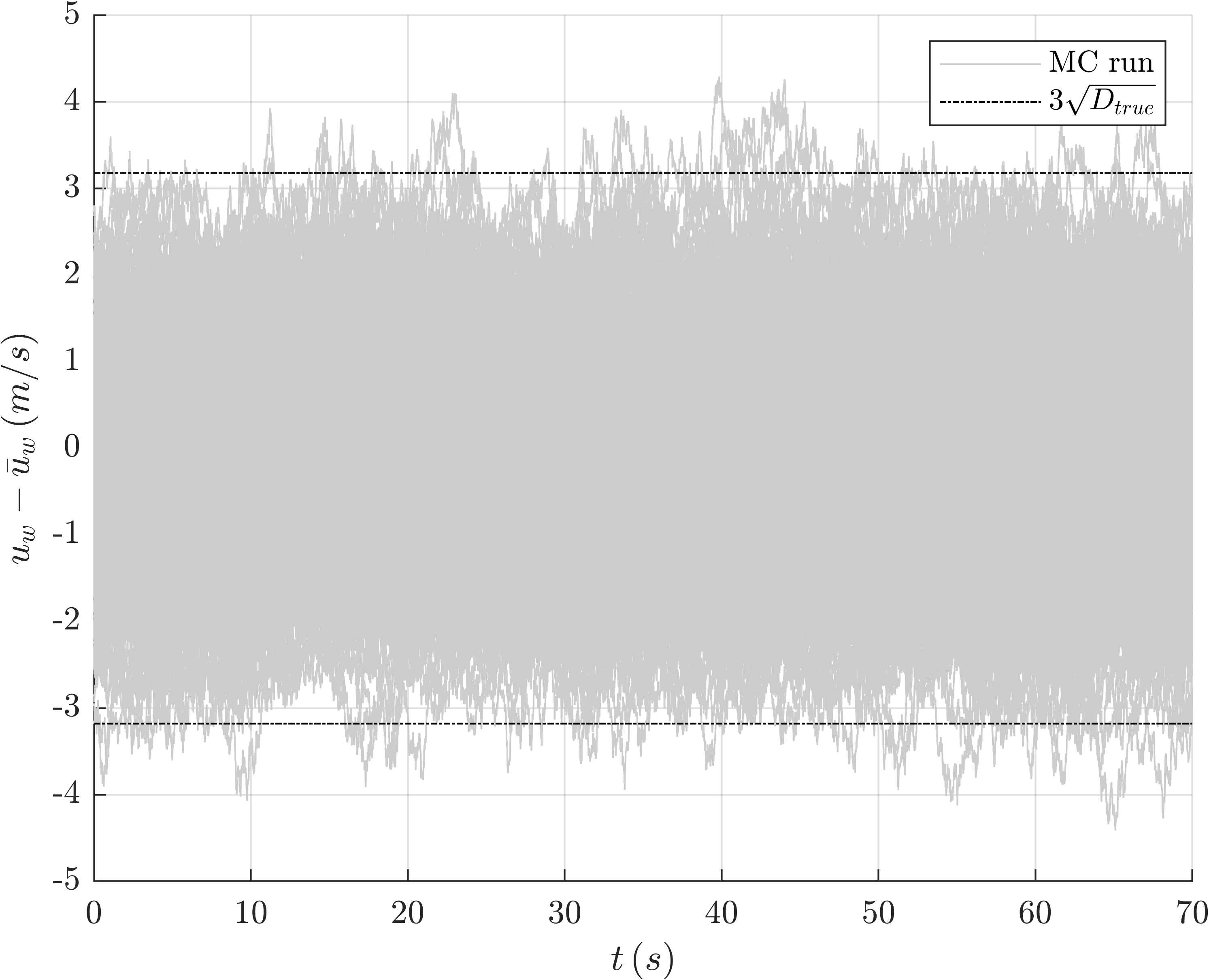}
\caption{Wind gust dispersions\label{fig:wind_gust_dispersion}}
\end{figure}

\begin{figure}[hbt!]
\centering
\includegraphics[width=3.0in]{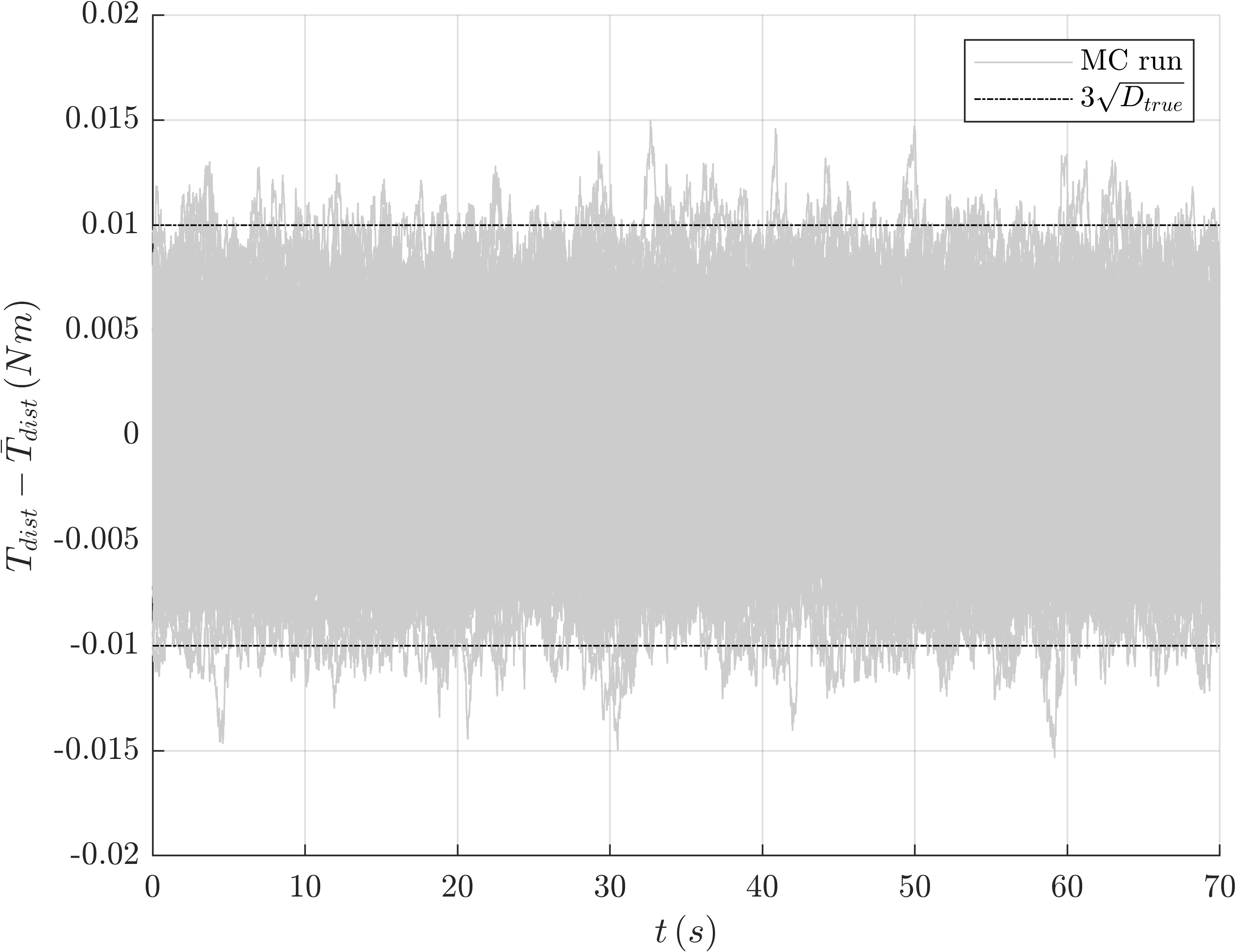}
\caption{Disturbance torque dispersions\label{fig:disturbance_torque}}
\end{figure}

Figures \ref{fig:Position-dispersion-north} and  \ref{fig:Position-dispersion-east} illustrate the position dispersions in the North and East directions.  Figures \ref{fig:Position-dispersion-along-track} and \ref{fig:Position-dispersion-cross-track} show similar results but in an Along-Track/Cross-Track coordinate system. The LinCov simulation provides a consistent estimate of vehicle position dispersions, with only occasional outliers occurring upon exiting the GPS-denied region.  This is due to the large position dispersions that accrue while inside the GPS-denied region, which cause a variation in the timing of GPS measurement availability.  The result is that a small portion of the Monte Carlo runs exit the GPS-denied region late and therefore receive the first GPS measurement later than the majority of runs.  Variability in the timing of events, or event triggers, is outside the scope of this research and has been considered in~\cite{geller_event_2009}.  Figures \ref{fig:velocity_dispersion} to \ref{fig:disturbance_torque} show similar results for the remaining truth states.  \revision{The truth state dispersions in Figs.~\ref{fig:Position-dispersion-north} to~\ref{fig:disturbance_torque} contain the coupled effect of the GNC algorithms with the exogenous wind and torque disturbances $u_w$ and $T_{dist}$, continuous measurement noise $\eta_a$ and $\eta_\omega$, and discrete measurement noise $\nu_{n}$, $\nu_{e}$, and $\nu_{d}$.  The results demonstrate that despite the nonlinear nature of the system the CL-LinCov architecture accurately captures the statistics of the truth state dispersions.}

\begin{figure}[hbt!]
\centering
\includegraphics[width=3.0in]{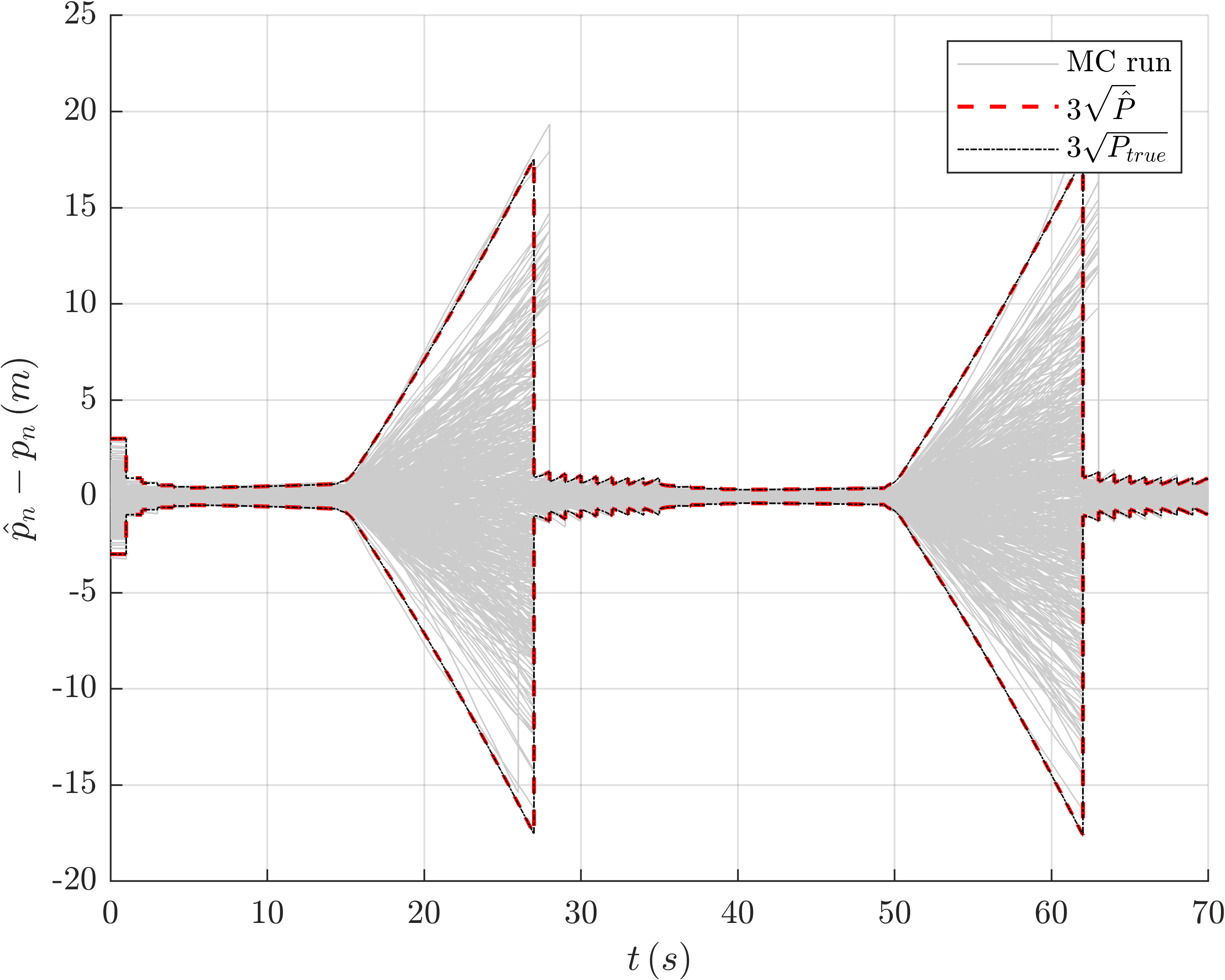}
\caption{North position estimation errors\label{fig:esterror_north}}
\end{figure}

\begin{figure}[hbt!]
\centering
\includegraphics[width=3.0in]{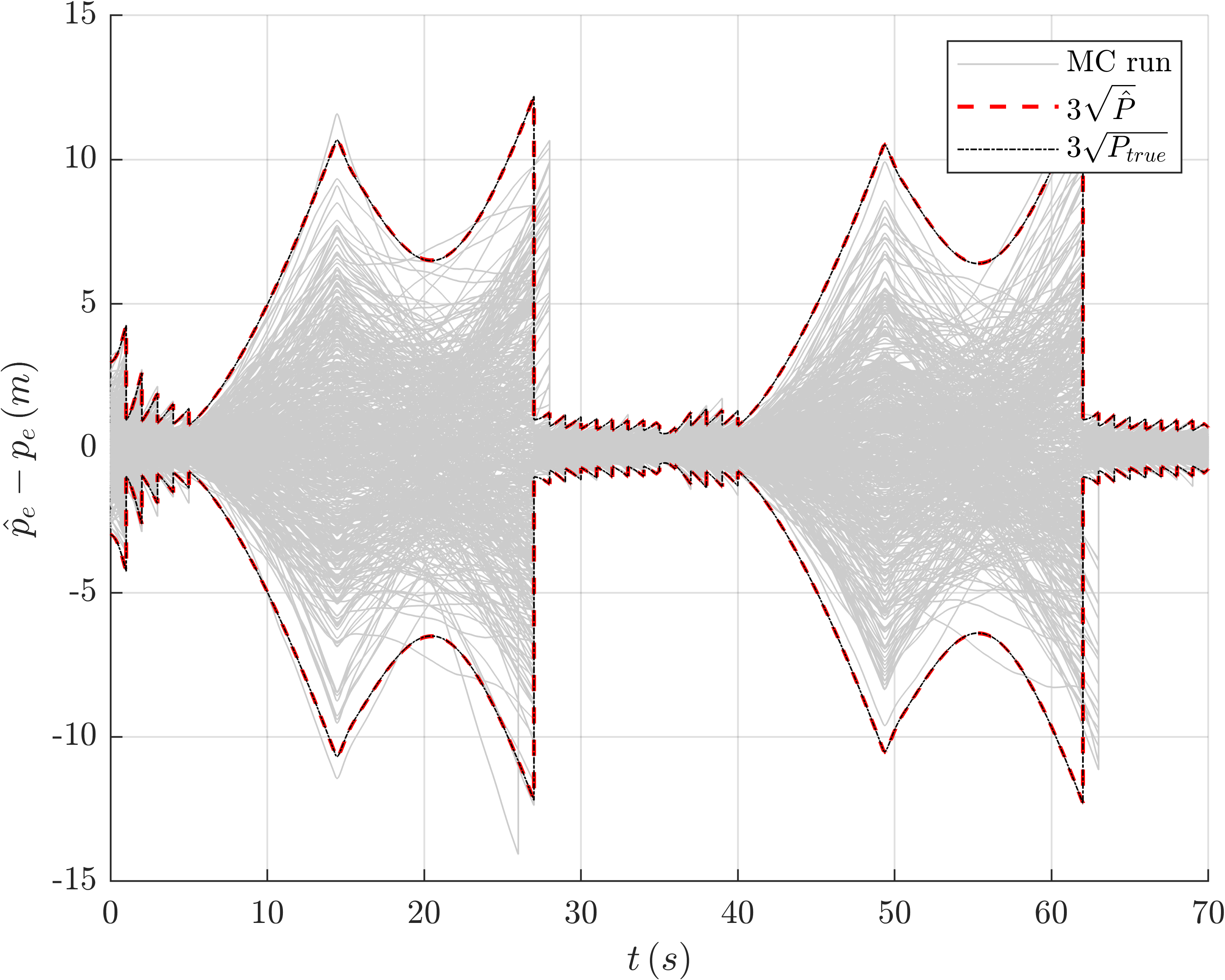}
\caption{East position estimation errors\label{fig:esterror_east}}
\end{figure}

\begin{figure}[hbt!]
\centering
\includegraphics[width=3.0in]{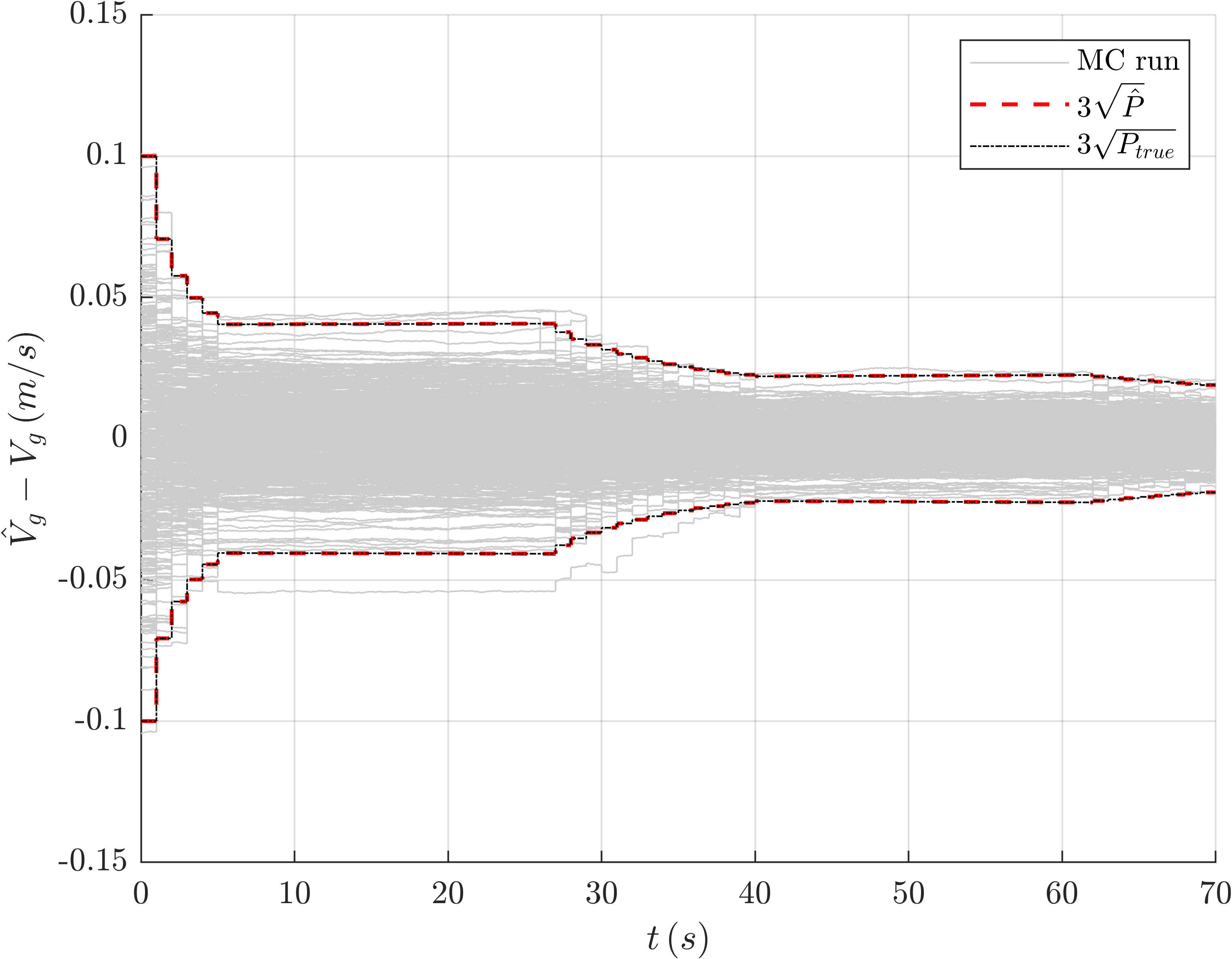}
\caption{Ground speed estimation errors\label{fig:esterror_v}}
\end{figure}

\begin{figure}[hbt!]
\centering
\includegraphics[width=3.0in]{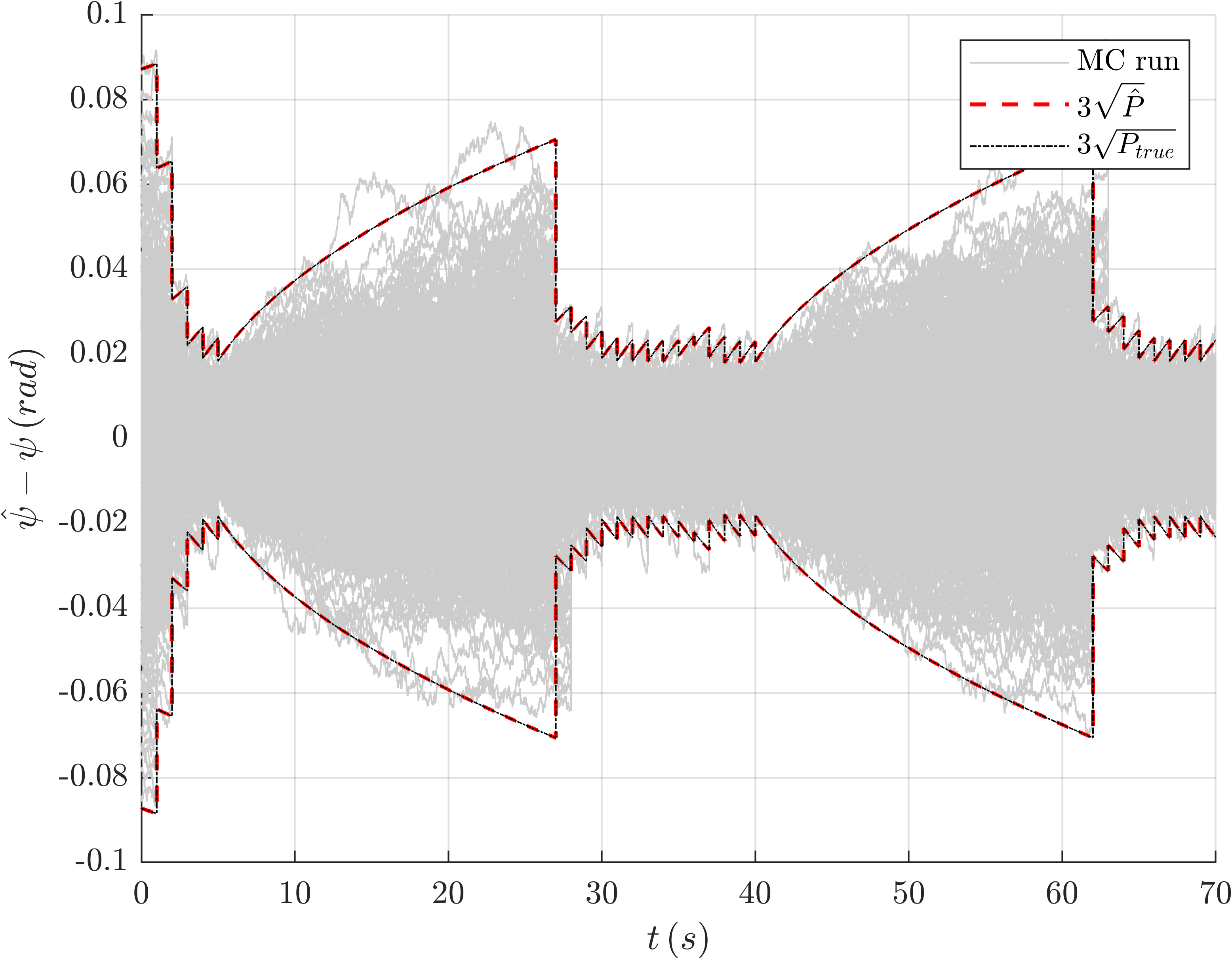}
\caption{Heading estimation errors\label{fig:esterror_psi}}
\end{figure}

Figures \ref{fig:esterror_north} to \ref{fig:esterror_psi} illustrate the estimation errors from the Monte Carlo simulations, the 3$\sigma$ standard deviation from the EKF, and the 3$\sigma$ standard deviation derived from the augmented covariance matrix. Agreement is observed for all navigation states, suggesting that the EKF is consistent with the truth models in the Monte Carlo simulation and the CL-LinCov framework.

This sections illustrates that the developed CL-LinCov framework is consistent with the Monte Carlo simulation.  CL-LinCov produces vehicle dispersion covariances that accurately predict the random variations of the vehicle from the planned path. System analyses, trade studies, sensitivities studies, or, as in the case of this research, path planning can therefore utilize the computed covariances in place of Monte Carlo methods.

\subsection{Path Planning Demonstration\label{subsec:Path-Planning-Demonstration}}
\begin{figure}[hbt!]
\centering
\includegraphics[width=3.5in]{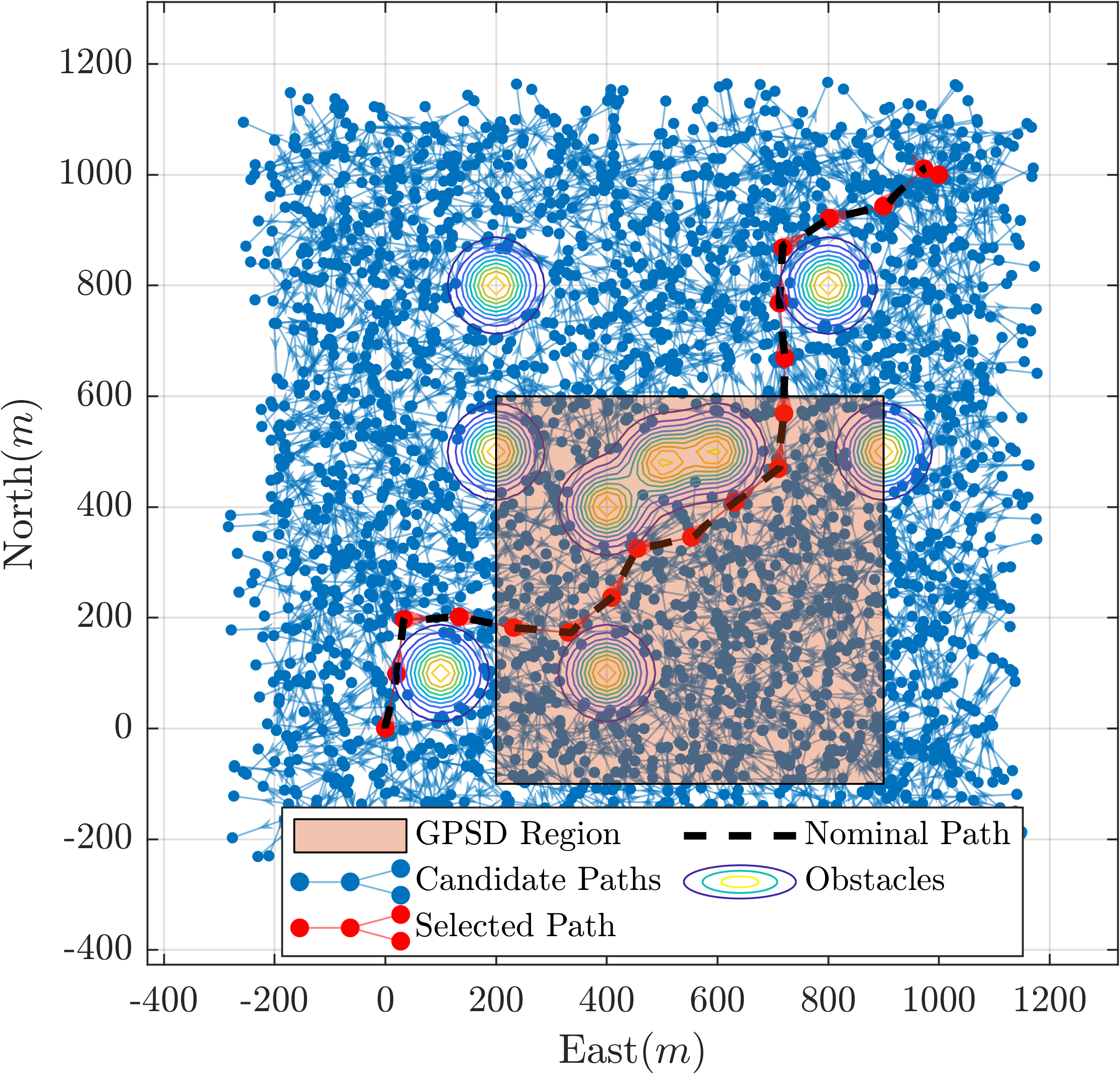}
\caption{Path planning Scenario 1: Collision threshold = 0.01\label{fig:pthresh_0_01}}
\end{figure}

\begin{figure}[hbt!]
\centering
\includegraphics[width=3.5in]{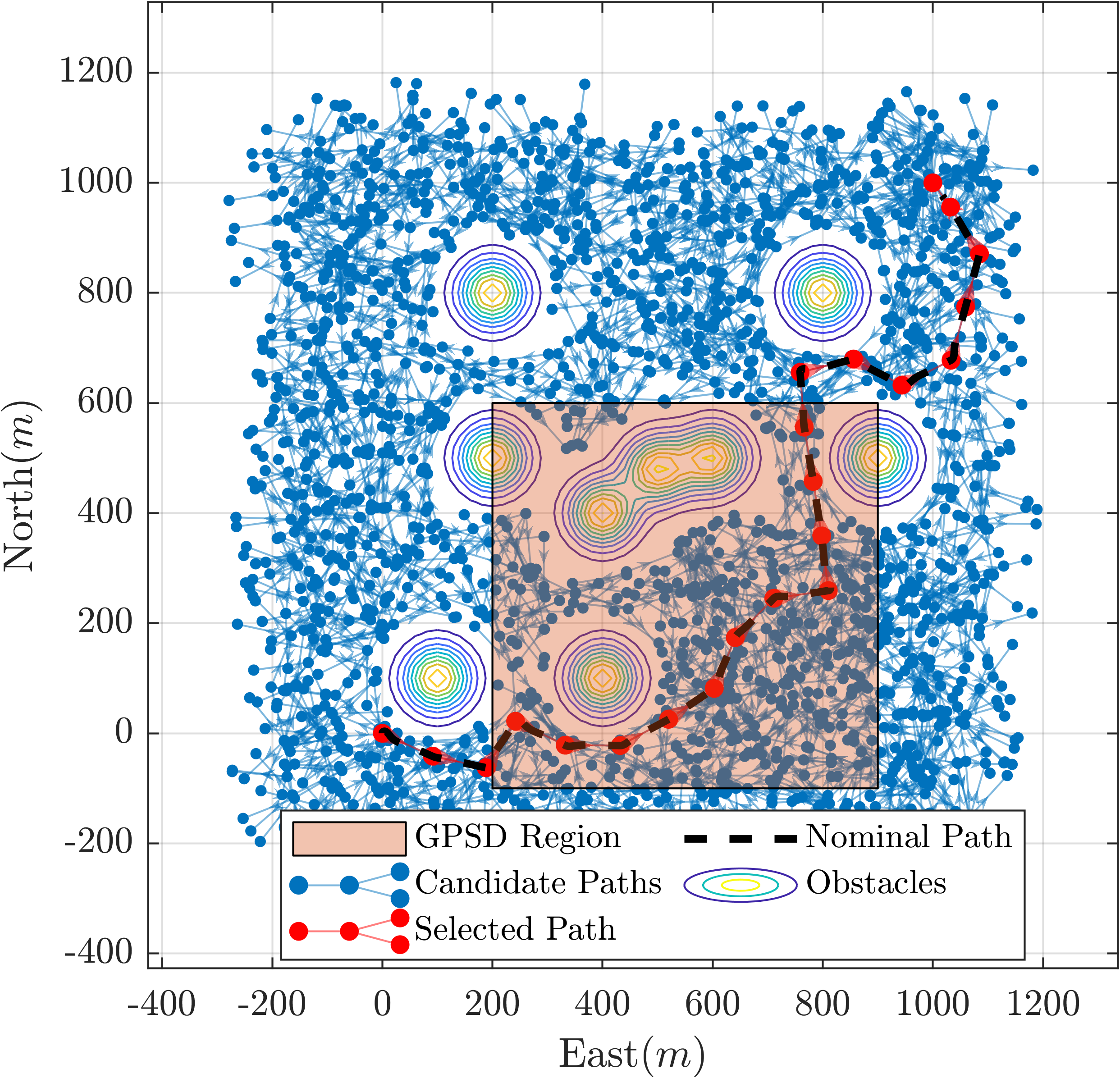}
\caption{Path planning Scenario 2: Collision threshold = 0.001\label{fig:pthresh_0_001}}
\end{figure}

\begin{figure}[hbt!]
\centering
\includegraphics[width=3.5in]{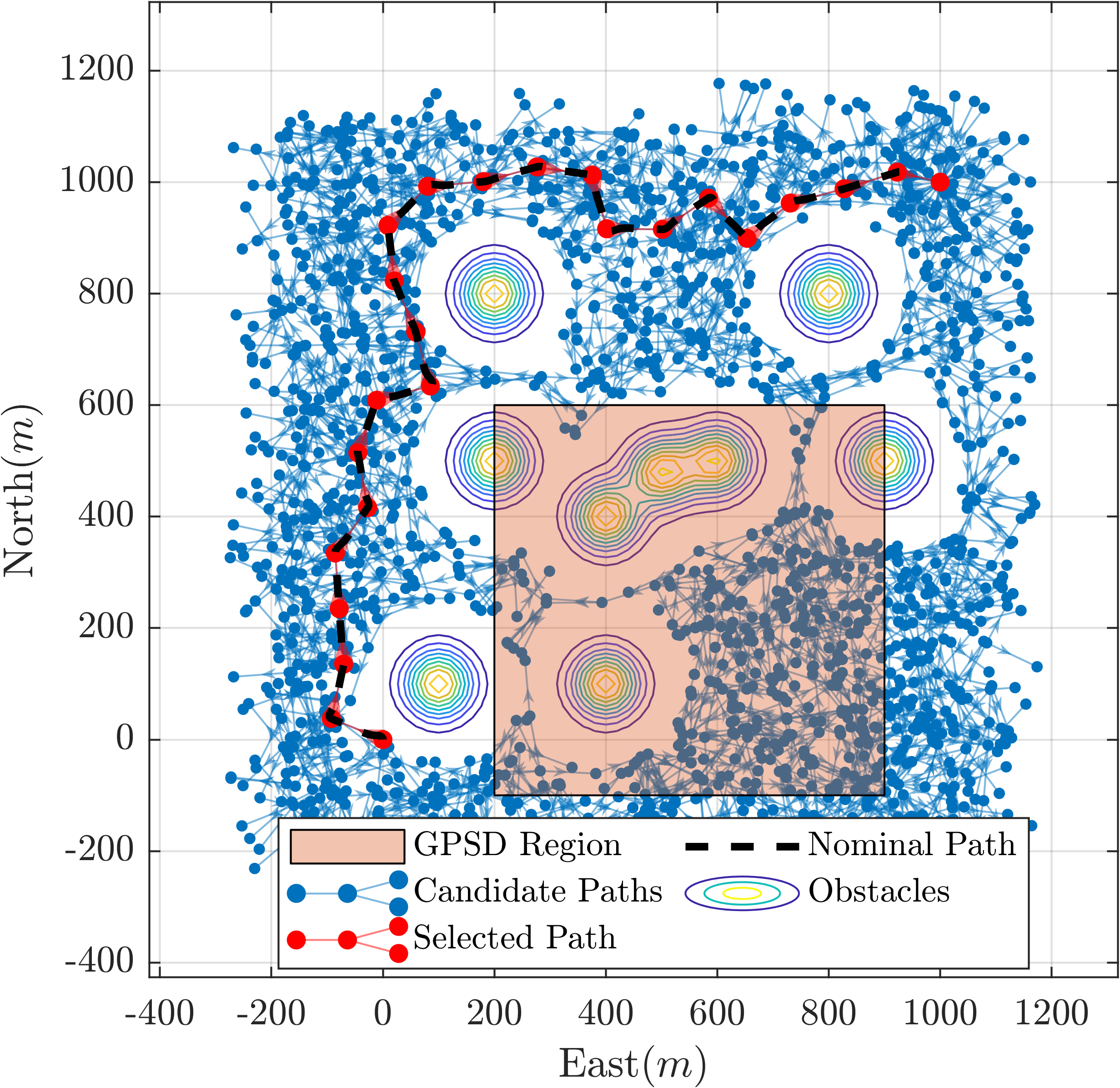}
\caption{Path planning Scenario 3: Collision threshold = 0.0001\label{fig:pthresh_0_0001}}
\end{figure}

This section provides an application of the CL-LinCov framework in the RRT path planner for a \revision{static,} uncertain obstacle field. Figures \ref{fig:pthresh_0_01}, \ref{fig:pthresh_0_001}, and \ref{fig:pthresh_0_0001} demonstrate the results of a 3000-iteration RRT path planning simulation for three scenarios, where the probability of collision threshold varies from 0.01, 0.001, and 0.0001, respectively.  For all scenarios, each obstacle has a location uncertainty of 40 meters $1\sigma$, with a \revision{bounding box size $\boldsymbol{l}=10$} meters.  The UAV begins at position [0,0] and is tasked with planning a path to the destination at position [1000,1000]. The blue dots and lines correspond to vertices and edges in the RRT graph. The multi-colored circles correspond to the contours of the Gaussian PDF, which defines the location of the obstacle. The red dots and lines correspond to the path in the RRT with minimum travel time, i.e. the selected path. The black dashed line represents the nominal reference trajectory, about which the position dispersions are computed. Finally, the shaded region corresponds to an area of GPS denial, where position and ground speed measurements are not available, causing the vehicle position disperions to increase over time.

\begin{figure}[hbt!]
\centering
\includegraphics[width=3.0in]{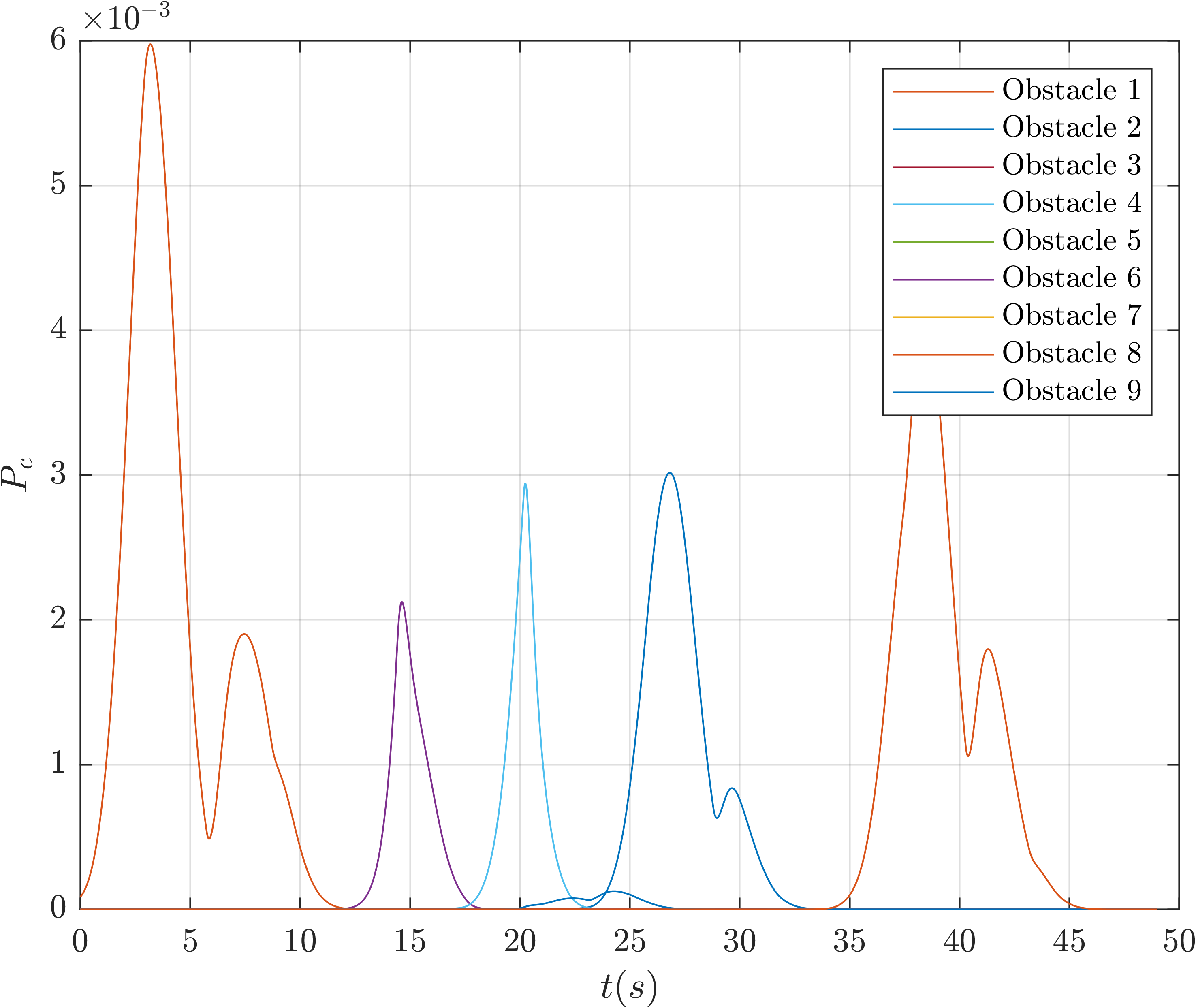}
\caption{Collision probabilities for each obstacle, along the lowest-cost path for Scenario 1: Collision threshold = 0.01\label{fig:pthresh_0_01_prob_collision}}
\end{figure}

\begin{figure}[hbt!]
\centering
\includegraphics[width=3.0in]{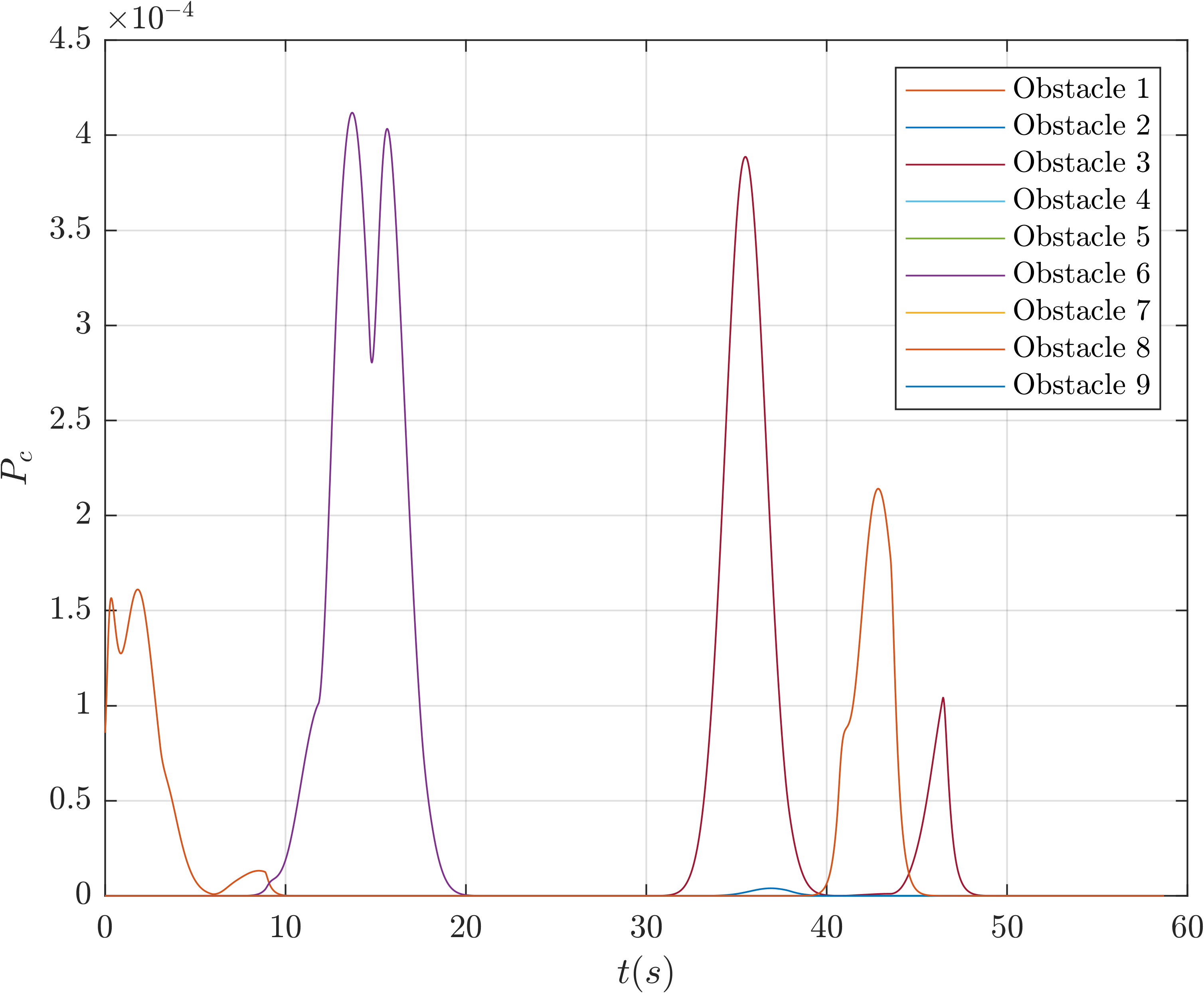}
\caption{Collision probabilities for each obstacle, along the lowest-cost path for Scenario 2: Collision threshold = 0.001\label{fig:pthresh_0_001_prob_collision}}
\end{figure}

\begin{figure}[hbt!]
\centering
\includegraphics[width=3.0in]{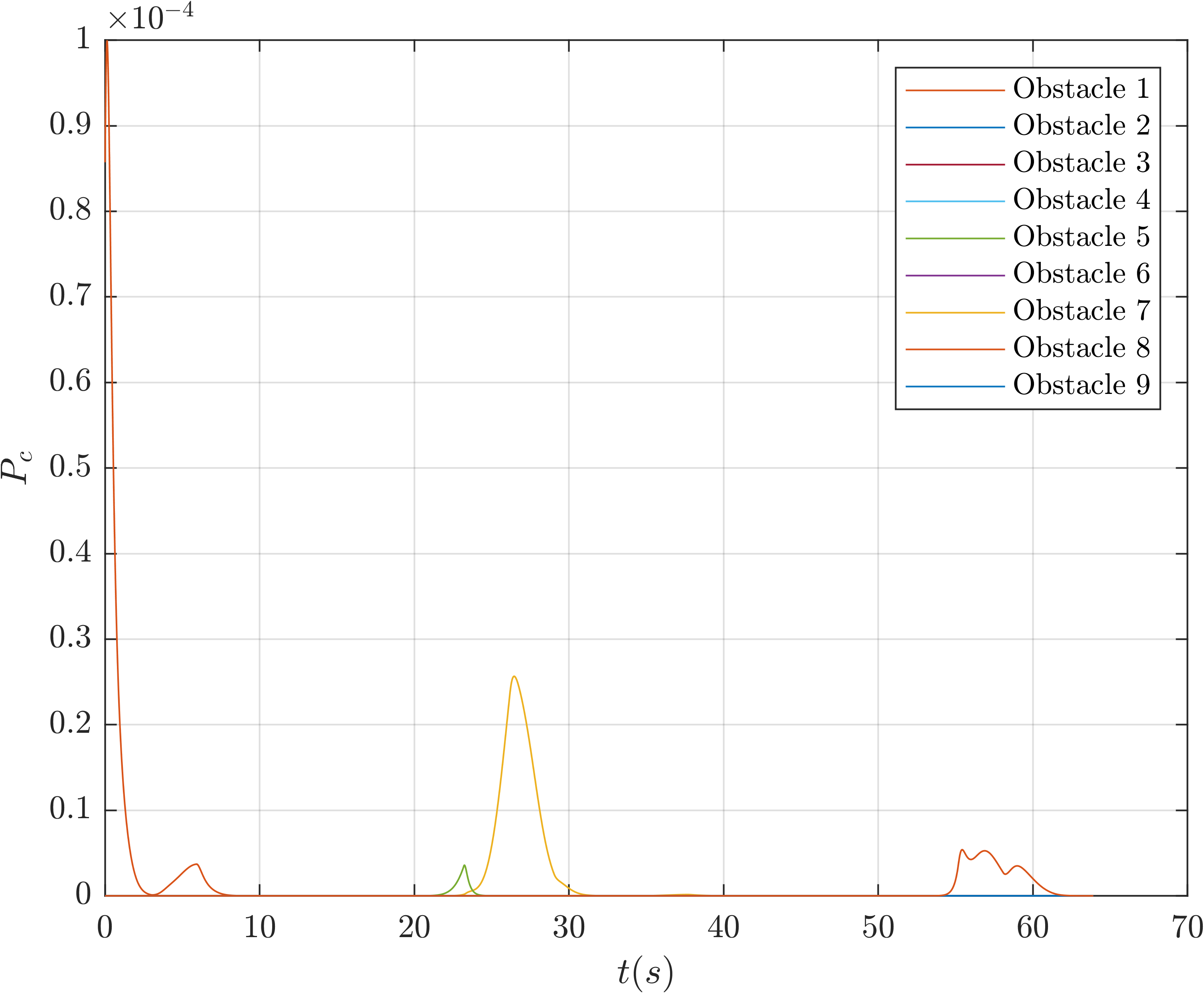}
\caption{Collision probabilities for each obstacle, along the lowest-cost path for Scenario 3: Collision threshold = 0.0001\label{fig:pthresh_0_0001_prob_collision}}
\end{figure}

As designed, the RRT algorithm, augmented by knowledge of the UAV position dispersion covariance and obstacle covariance, discovers paths that meet the specified probability of collision threshold.  For the large 0.01 threshold value of Fig. \ref{fig:pthresh_0_01}, the path planner considers many possible routes internal to the obstacle field, resulting in a more direct path to the destination.  As the probability of collision threshold decreases to 0.001 in Fig. \ref{fig:pthresh_0_001}, the path planner becomes more conservative, only considering paths farther away from the uncertain obstacles.  In Fig. \ref{fig:pthresh_0_0001}, the path planner has become so conservative that very few candidate paths exist inside the obstacle field, resulting in the vehicle traveling around the exterior of the obstacles. In all cases, the planned path conforms to the specified probability of collision as verified by Figs. \ref{fig:pthresh_0_01_prob_collision} to \ref{fig:pthresh_0_0001_prob_collision}.

\section{Conclusion\label{sec:Conclusion}}
The ability to quickly plan safe paths in uncertain environments is an enabling technology for wide-spread adoption of autonomous vehicles. This manuscript develops a new closed-loop linear covariance (CL-LinCov) framework that directly computes the dispersion covariance of the vehicle about a candidate path. The framework linearizes the dynamics, guidance, navigation and control equations about the candidate path, accounting for the effects of measurement noise and availability as well as exogenous disturbances.  The framework is designed to accommodate a wide range of system architectures typical of autonomous systems, including cascaded guidance and control algorithms, output feedback controllers, and multi-use continuous sensors.  The framework is applied to a simplified model of a UAV in the horizontal plane, with straight-line following between way-points, and aided by GPS position and velocity measurements. The consistency of the CL-LinCov results is validated via Monte Carlo analysis for a scenario with multiple way-points both inside and outside of a GPS-denied region.	As a demonstration of its utility, the CL-LinCov framework is integrated with the rapidly-exploring random tree (RRT) path planning algorithm in the presence of uncertain obstacles.  The vehicle dispersion covariance is combined with obstacle location uncertainty to quantify the probability of collision about each candidate path. The CL-LinCov + RRT algorithm is shown to identify feasible paths to the destination, while maintaining the probability of collision below user-specified levels.

\section*{Appendix\label{sec:Appendix}}

\subsection{Linear Covariance Model Derivation\label{subsec:Linear-Covariance-Model}}

This appendix documents the linearization of the models in Section
\ref{subsec:Nonlinear-Truth-Model} to yield the LinCov framework
of Section \ref{subsec:Linear-Modeling}. Linearization of the truth
state dynamics Eq.~\eqref{eq:nonlin1}, continuous measurements Eq.~\eqref{eq:nonlin2},
guidance law Eq.~\eqref{eq:nonlin12}, and controller output Eq.~\eqref{eq:nonlin14}
yields the following expressions:
\begin{equation}
\delta\dot{\boldsymbol{x}}=F_{x}\delta\boldsymbol{x}+F_{u}\delta\boldsymbol{u}+B\boldsymbol{w}\label{eq:lin4}
\end{equation}
\begin{equation}
\delta\tilde{\boldsymbol{y}}=C_{x}\delta\boldsymbol{x}+C_{u}\delta\boldsymbol{u}+\boldsymbol{\eta}\label{eq:lin7-1}
\end{equation}
\begin{equation}
\delta\hat{\boldsymbol{x}}^{*}=N_{\hat{x}}\delta\hat{\boldsymbol{x}}+N_{\tilde{y}}\delta\tilde{\boldsymbol{y}}\label{eq:lin6}
\end{equation}
\begin{equation}
\delta\boldsymbol{u}=G_{\check{x}}\delta\check{\boldsymbol{x}}+G_{\hat{x}}\delta\hat{\boldsymbol{x}}+G_{\hat{x}^{*}}\delta\hat{\boldsymbol{x}}^{*}+G_{\tilde{y}}\delta\tilde{\boldsymbol{y}}\label{eq:lin5}
\end{equation}
Substitution of Eq.~\eqref{eq:lin7-1} and Eq.~\eqref{eq:lin6} into Eq.~\eqref{eq:lin5}
yields
\begin{eqnarray}
\delta\boldsymbol{u} & = & \left(G_{\hat{x}^{*}}N_{\tilde{y}}C_{x}+G_{\tilde{y}}C_{x}\right)\delta\boldsymbol{x}\nonumber \\
 & + & \left(G_{\hat{x}}+G_{\hat{x}^{*}}N_{\hat{x}}\right)\delta\hat{\boldsymbol{x}}+G_{\check{x}}\delta\check{\boldsymbol{x}}\nonumber \\
 & + & \left(G_{\hat{x}^{*}}N_{\tilde{y}}C_{u}+G_{\tilde{y}}C_{u}\right)\delta\boldsymbol{u}\nonumber \\
 & + & \left(G_{\hat{x}^{*}}N_{\tilde{y}}+G_{\tilde{y}}\right)\boldsymbol{\eta}
\end{eqnarray}
Isolation of the control dispersion term yields
\begin{eqnarray}
\delta\boldsymbol{u} & = & S\left(G_{\hat{x}^{*}}N_{\tilde{y}}C_{x}+G_{\tilde{y}}C_{x}\right)\delta\boldsymbol{x}\nonumber \\
 & + & S\left(G_{\hat{x}}+G_{\hat{x}^{*}}N_{\hat{x}}\right)\delta\hat{\boldsymbol{x}}\nonumber \\
 & + & SG_{\check{x}}\delta\check{\boldsymbol{x}}+S\left(G_{\hat{x}^{*}}N_{\tilde{y}}+G_{\tilde{y}}\right)\boldsymbol{\eta}\label{eq:lin-delu}
\end{eqnarray}
where
\begin{equation}\label{eq:S-matrix}
S=\left(I_{n_{u}\times n_{u}}-G_{\hat{x}^{*}}N_{\tilde{y}}C_{u}-G_{\tilde{y}}C_{u}\right)^{-1}
\end{equation}
Substitution of Eq.~\eqref{eq:lin-delu} into Eq.~\eqref{eq:lin4} yields the
final form of the linearized truth state dispersion dynamics:
\begin{eqnarray}
\delta\dot{\boldsymbol{x}} & = & \left[F_{x}+F_{u}S\left(G_{\hat{x}^{*}}N_{\tilde{y}}C_{x}+G_{\tilde{y}}C_{x}\right)\right]\delta\boldsymbol{x}\nonumber \\
 & + & F_{u}S\left(G_{\hat{x}}+G_{\hat{x}^{*}}N_{\hat{x}}\right)\delta\hat{\boldsymbol{x}}\nonumber \\
 & + & F_{u}SG_{\check{x}}\delta\check{\boldsymbol{x}}\nonumber \\
 & + & F_{u}S\left(G_{\hat{x}^{*}}N_{\tilde{y}}+G_{\tilde{y}}\right)\boldsymbol{\eta}+B\boldsymbol{w}
\end{eqnarray}

Linearization of the navigation state dynamics Eq.~\eqref{eq:nonlin5} yields
\begin{equation}
\delta\dot{\hat{\boldsymbol{x}}}=\hat{F}_{\hat{x}}\delta\hat{\boldsymbol{x}}+\hat{F}_{\tilde{y}}\delta\tilde{\boldsymbol{y}}\label{eq:lin16}
\end{equation}
Substitution of Eq.~\eqref{eq:lin6} and Eq.~\eqref{eq:lin5} into Eq.~\eqref{eq:lin7-1}
yields
\begin{eqnarray}
\delta\tilde{\boldsymbol{y}} & = & C_{x}\delta\boldsymbol{x}\nonumber \\
 & + & \left(C_{u}G_{\hat{x}}+C_{u}G_{\hat{x}^{*}}N_{\hat{x}}\right)\delta\hat{\boldsymbol{x}}\nonumber \\
 & + & C_{u}G_{\check{x}}\delta\check{\boldsymbol{x}}\nonumber \\
 & + & \left(C_{u}G_{\hat{x}^{*}}N_{\tilde{y}}+C_{u}G_{\tilde{y}}\right)\delta\tilde{\boldsymbol{y}}+\boldsymbol{\eta}
\end{eqnarray}
Isolation of the measurement dispersion term yields
\begin{eqnarray}
\delta\tilde{\boldsymbol{y}} & = & TC_{x}\delta\boldsymbol{x}+T\left(C_{u}G_{\hat{x}}+C_{u}G_{\hat{x}^{*}}N_{\hat{x}}\right)\delta\hat{\boldsymbol{x}}\nonumber \\
 & + & TC_{u}G_{\check{x}}\delta\check{\boldsymbol{x}}+T\boldsymbol{\eta}\label{eq:lin-dely}
\end{eqnarray}
where
\begin{equation}\label{eq:T-matrix}
T=\left(I_{n_{y}\times n_{y}}-C_{u}G_{\hat{x}^{*}}N_{\tilde{y}}-C_{u}G_{\tilde{y}}\right)^{-1}
\end{equation}
Substitution of Eq.~\eqref{eq:lin-dely} into Eq.~\eqref{eq:lin16} yields the
final form of the linearized navigation state dispersions
\begin{eqnarray}
\delta\dot{\hat{\boldsymbol{x}}} & = & \hat{F}_{\tilde{y}}TC_{x}\delta\boldsymbol{x}\\
& + & \left(\hat{F}_{\hat{x}}+\hat{F}_{\tilde{y}}TC_{u}G_{\hat{x}}+\hat{F}_{\tilde{y}}TC_{u}G_{\hat{x}^{*}}N_{\hat{x}}\right)\delta\hat{\boldsymbol{x}}\\
 & + & \hat{F}_{\tilde{y}}TC_{u}G_{\check{x}}\delta\check{\boldsymbol{x}}+\hat{F}_{\tilde{y}}T\boldsymbol{\eta}
\end{eqnarray}
\revision{Note that the inverse operation in Eqs.~\ref{eq:S-matrix} and~\ref{eq:T-matrix} have the potential to become ill-conditioned. In the authors' experience, the scenarios where this occurs in practice are uncommon. Care must be taken, nevertheless, to ensure invertability for a specific application.}

Linearizing the controller state dynamics Eq.~\eqref{eq:nonlin13} yields
\begin{equation}
\delta\dot{\check{\boldsymbol{x}}}=\check{F}_{\hat{x}}\delta\hat{\boldsymbol{x}}+\check{F}_{\hat{x}^{*}}\delta\hat{\boldsymbol{x}}^{*}\label{eq:lin-xcheckdot}
\end{equation}
Substitution of Eq.~\eqref{eq:lin6} into Eq.~\eqref{eq:lin-xcheckdot} yields 
\begin{equation}
\delta\dot{\check{\boldsymbol{x}}}=\check{F}_{\hat{x}}\delta\hat{\boldsymbol{x}}+\check{F}_{\hat{x}^{*}}\left(N_{\hat{x}}\delta\hat{\boldsymbol{x}}+N_{\tilde{y}}\delta\tilde{\boldsymbol{y}}\right)
\end{equation}
Substitution of Eq.~\eqref{eq:lin-dely} yields the final form of the controller state dispersion dynamics:
\begin{eqnarray}
\delta\dot{\check{\boldsymbol{x}}} & = & \check{F}_{\hat{x}^{*}}N_{\tilde{y}}TC_{x}\delta\boldsymbol{x}\nonumber \\
 & + & \left[\check{F}_{\hat{x}}+\check{F}_{\hat{x}^{*}}N_{\hat{x}}+\check{F}_{\hat{x}^{*}}N_{\tilde{y}}T\left(C_{u}G_{\hat{x}}+C_{u}G_{\hat{x}^{*}}N_{\hat{x}}\right)\right]\delta\hat{\boldsymbol{x}}\nonumber \\
 & + & \check{F}_{\hat{x}^{*}}N_{\tilde{y}}TC_{u}G_{\check{x}}\delta\check{\boldsymbol{x}}\nonumber \\
 & + & \check{F}_{\hat{x}^{*}}N_{\tilde{y}}T\boldsymbol{\eta}
\end{eqnarray}

The remainder of this derivation linearizes the effect of the Kalman update, which for the controller and truth state dispersions has no effect. Therefore,
\begin{equation}
\delta\boldsymbol{x}_{k}^{+}=\delta\boldsymbol{x}_{k}^{-}
\end{equation}
\begin{equation}
\delta\check{\boldsymbol{x}}_{k}^{+}=\delta\check{\boldsymbol{x}}_{k}^{-}
\end{equation}
Linearizing the navigation state update Eqs.~\eqref{eq:nonlin4},
\eqref{eq:nonlin6}, and Eq.~\eqref{eq:nonlin7} yields
\begin{equation}
\delta\hat{\boldsymbol{x}}_{k}^{+}=\delta\hat{\boldsymbol{x}}_{k}^{-}+\hat{K}_{k}\left[\delta\tilde{\boldsymbol{z}}_{k}-\delta\hat{\tilde{\boldsymbol{z}}}_{k}\right]
\end{equation}
where
\begin{equation}
\delta\tilde{\boldsymbol{z}}_{k}=H_{x}\delta\boldsymbol{x}_{k}^{-}+\boldsymbol{\nu}_{k}
\end{equation}
\begin{equation}
\delta\hat{\tilde{\boldsymbol{z}}}_{k}=\hat{H}_{\hat{x}}\delta\hat{\boldsymbol{x}}_{k}^{-}
\end{equation}
Substitution yields the final form of the navigation state dispersion
update
\begin{equation}
\delta\hat{\boldsymbol{x}}_{k}^{+}=\left(I_{\hat{n}\times\hat{n}}-\hat{K}_{k}\hat{H}_{\hat{x}}\right)\delta\hat{\boldsymbol{x}}_{k}^{-}+\hat{K}_{k}H_{x}\delta\boldsymbol{x}_{k}^{-}+\hat{K}_{k}\boldsymbol{\nu}_{k}
\end{equation}

\section*{Funding Sources}
This research was funded in part by the U.S. Air Force Research Lab Summer Faculty Fellowship Program.

\bibliography{lincov_path_planning}

\end{document}